\documentclass[english]{article}
\usepackage{jheppub}
\usepackage[utf8]{inputenc}
\usepackage[english]{babel}
\usepackage{amsmath}
\usepackage{amsthm}
\usepackage{amssymb}
\usepackage{babel}
\usepackage{caption}
\usepackage{subcaption}
\usepackage{comment}
\usepackage[multiple]{footmisc}

\usepackage[noend]{algpseudocode}
\usepackage{algorithm,algorithmicx,algpseudocode}
\newcommand{\bb}[1]{\boldsymbol{#1}}

\newcommand{\mE}{\mathbb E}

\newcommand{\red}[1]{\textcolor{black}{#1}}

\def\lc{\left\lceil}   
\def\rc{\right\rceil}
\newenvironment{algocolor}{%
   \setlength{\parindent}{0pt}
   \itshape
   \color{black}
}{}
\usepackage{xcolor}
\begin{document}
\title{Regularization, early-stopping and dreaming: a Hopfield-like setup to address generalization and overfitting}
\author[a,1]{E. Agliari,}
\author[b,1]{F. Alemanno,}
\author[a,1]{M. Aquaro,}
\author[a,1]{A. Fachechi}
\affiliation[a]{Dipartimento di Matematica ``Guido Castelnuovo'', Sapienza Universit\`a di Roma, Italy}
\affiliation[b]{Dipartimento di Matematica, Universit\`a di Bologna, Italy}
\affiliation[1]{GNFM-INdAM, Gruppo Nazionale di Fisica Matematica (Istituto Nazionale di Alta Matematica), Italy}
\emailAdd{agliari@mat.uniroma1.it}

\abstract{In this work we approach attractor neural networks from a machine learning perspective: we look for optimal network parameters by applying a gradient descent over a regularized loss function. Within this framework, the optimal neuron-interaction matrices turn out to be a class of matrices which correspond to Hebbian kernels revised by a reiterated unlearning protocol. Remarkably, the extent of such unlearning is proved to be related to the regularization hyperparameter of the loss function and to the training time. Thus, we can design strategies to avoid overfitting that are formulated in terms of regularization and early-stopping tuning. The generalization capabilities of these attractor networks are also investigated: analytical results are obtained for random synthetic datasets, next, the emerging picture is corroborated by numerical experiments that highlight the existence of several regimes (i.e., overfitting, failure and success) as the dataset parameters are varied.}

\maketitle

\section{Introduction}

The Hopfield model is probably the best-known example of attractor neural network \cite{Amari-1972, Little-1974, Hopfield}: this is constituted by a set of $N$ binary units, referred to as neurons, that interact pairwise and whose state is iteratively updated by a non-linear activation function, in such a way that the new state of neuron $i$ depends on the signal acting on $i$ and stemming from the neighbouring neurons. 
A suitable choice of the neuron interaction matrix, {denoted as $\bb J \in \mathbb R^{N \times N}$}, can ensure the attractivity of a number of patterns, that we want to store and retrieve. More precisely, one initializes the neural configuration by setting the values of the units close to a pattern $\bb \xi^1 \in \{-1, +1\}^N$, {this configuration represents the input supplied to the machine and may correspond to a corrupted version of $\bb \xi^1$; repeated updates of neurons are then performed until convergence to a fixed point and, if this matches $\bb \xi^1$,} this state is interpreted as the retrieval of the information codified by $\bb \xi^1$. {The same is expected to occur for any target pattern, say $\bb \xi^{\mu}$, with $\mu=1,...,P$.}  

The popularity of the Hopfield model is also due to the fact that it is feasible of an analytical treatment and, in particular, it can be recognized as a spin-glass, {in such a way} that it can benefit from a broad collection of techniques developed to address disordered systems. Indeed, in the last decades, the model has been intensively investigated, and countless variations on theme have also been accounted for\footnote{{These include, for instance, the presence of biases or vacancies in pattern entries (see e.g., \cite{Sompo-Biased,ALM-AMC2022,ABGGM-PRL2012}), generalizations of the Hebbian kernel to include temporal correlations among patterns (see e.g., \cite{Cugliandolo-1993,ABDG-NN2013,AFM-JMP2020}) or synaptic noise (see e.g., \cite{AABCF-PRL2020,AD-EPJP2020,CCM-2022}), non-trivial underlying architectures (see e.g., \cite{WC-JPA2003,AABCT-JPA2013b,AMT-JSP2018}), or high-order interactions (see e.g., \cite{Baldi-PRL1987,HopKro1,AABF-NN2020}). Although these aspects are not directly related to the current work, it is worth stressing the long-standing, broad interest attracted by the model, further, the current approach can be extended to include such features.}}. {Remarkably, a significant fraction of these works spotlighted the structure of the neural interaction matrix: in the standard Hopfield model this is based on the so-called Hebb's rule \cite{Hebb-1949} $\bb J = \frac{\bb \xi \cdot \bb \xi^T}{N}$ and suitable revisions of this rule can give rise to better performances of the model in terms of number of storable and retrievable patterns \cite{Amit}. A successful class of these revisions apply unlearning protocols (see e.g., \cite{Hopfield-1983,Kanter-Sompolinsky-1987,GFV-JPhys1989,FAV-JPhys1990,DYD-1991,Plakhov-1995,FachechiNN,Marinari-2019,MarinariRuoccoZamponi-2022,serricchio2023daydreaming}), whose aim is to impair the attractiveness of configurations that do not correspond to any of the stored patterns (the convergence to those configurations, sometimes referred to as spurious states, is interpreted as a mistake of the machine).} More recently, {Hebb's rule} has also been revised to make it closer to a learning algorithm: {the ``reality'' that we want to retrieve is now not accessible, instead, a corrupted sample, say $\{\bb \xi^{\mu,1}, \bb \xi^{\mu,2}, ... \}$ is available and used to build $\bb J$} (see e.g., \cite{Fontanari-1990,AABD-NN2022,AAKBA-EPL2023,negri2023storage}). {This way, the available sample of data can be interpreted as a training set and the Hopfiled model can be employed for generalization tasks.} The bridge between a retrieval scenario and a machine learning setting has also been strengthened by leveraging the equivalence between Hopfield model and Boltzmann machines (see e.g., \cite{Contucci,CMS-PRE2011,M-PRE2017,leonelli2021effective,AlbertIEEE,AM-2020}).
However, a full mapping allowing for the role of regularization parameters, the emergence of overfitting or underfitting phenomena is still under construction (see e.g., \cite{CTB-2023,BCMS-2022,AACF-2023,zamri2022weighted,VCMZ-2023,zamri2024modified}). 

In this paper, we {try to contributed in filling this gap, focusing} on an unsupervised reconstruction problem: the dataset is made of a set of items belonging to different classes (the label being veiled) and we introduce a loss function for the neuron interaction matrix. 
The solution of our problem corresponds to a Hebbian kernel subjected to a certain amount $t_d$ of unlearning iterations and we prove that $t_d$ is related to the regularization hyperparameter, which, in turn, can be related to the training time in the un-regularized version of the problem. 
This framework allows us to inspect the emergence of overfitting phenomena and therefore to {conceive recipes} for an optimal training time.
Specifically,  the system stores each pattern as a minimum of the Lyapunov function associated to the neural dynamics (this can be interpreted as a cost function or as an energy function); minima corresponding to the same class form a cluster, and -- when the number of examples per class is large enough -- these minima do coalesce into a single minimum.  In this scenario, there emerge both intra-class and inter-class correlations and we find that the role of $t_d$ (or, equivalently, of the regularization or of the training time) is to disentangle such correlations starting from the lowest ones: as $t_d$ is increased, the minima corresponding to different classes are shifted, their overlap is reduced and the system gets able to generalize from examples; by further increasing $t_d$, minima corresponding to patterns belonging to the same classes get shifted too and the system starts to overfit.

In what follows, we detail these results by first introducing the loss function associated with our problem and showing that the neural interaction matrix that minimizes the loss function corresponds to the revised Hebbian kernel studied in \cite{FachechiNN,AABF-JStat2019} (Sec.~\ref{sec:stability}). Subsequently, we find a relation between regularization, training time and unlearning time, and we present numerical experiments on structureless and structured datasets (Sec.~\ref{sec:dreaming}). These results constitute the premises for a thorough discussion on the emergence of overfitting phenomena (Sec.~\ref{sec:overfitting}) and a corroboration by numerics (Sec.~\ref{sec:numerics}). Lastly, we conclude our paper by offering a concise outlook and final remarks (Sec.~\ref{sec:conclusions}).
{Technical details are collected in the Appendices.}

\section{From the stability condition to the minimization problem\label{sec:stability}}
	
The basic principle underlying an attractor neural network is that each pattern making up the set $\bb \xi = \{\bb \xi^\mu\}_{\mu=1}^P$ and encoding relevant information is associated to an attracting fixed point for the network dynamics. We assume that patterns are $N$-dimensional binary vectors and that the network units $\bb \sigma= (\sigma_1,\dots, \sigma_N) \in \{-1, +1\}^N$ interact pairwise as specified by the symmetric\footnote{The symmetry constraint on the coupling matrix is traditionally adopted so that, in the statistical mechanical approach, detailed balance principle holds and this directly implies the stochastic relaxation to an equilibrium distribution in Boltzmann-Gibbs form, see e.g., \cite{Coolen}.} coupling matrix $\bb J$, then, we set up the network evolution as
\begin{equation}\label{eq:zeronoise}
	\bb \sigma (t+1)= \text{sign} (\bb \varphi(\boldsymbol \sigma (t)),
\end{equation}
where $\varphi_i (\boldsymbol \sigma (t)) \equiv \sum_{j\neq i} J_{ij}\sigma_j (t)$ represents the signal reaching unit $i$ at time $t$ and the sign function acts component-wise. This dynamics is applied sequentially and exhibits the following Lyapunov function (see e.g., \cite{Coolen})
\begin{equation}\label{eq:energy_function}
\mathcal E (\boldsymbol \sigma) = - \frac{1}{2} \sum_{i,j} \sigma_i J_{ij} \sigma_j. 
\end{equation}
In this work, we will retain {a deterministic}\footnote{Stochastic realizations of the dynamics \eqref{eq:zeronoise} work in similar ways, apart from the fact that the system is not fixed in the precise configuration given by the $\mu$-th pattern, but it is free to explore (at some level given by the {degree of stochasticity}) the associated attraction basin. This is strictly true if {two conditions hold:} $i.$ the initial configuration is close (in the Hamming sense) to the target pattern and $ii.$ in the thermodynamic limit, where ergodicity breaking occurs (in the finite-size case, the transition between different attractors is exponentially small in the system size but non-vanishing).} {evolution} and, for the moment, we take $\boldsymbol J$ as quenched and we exclude the presence of biases. In this framework, the stability condition for a retrieval configuration, e.g., $\boldsymbol \sigma (t) = \boldsymbol \xi^{\mu}$ without loss of generality, reads as
\begin{equation}\label{eq:stability}
\xi^\mu_i \varphi_i(\bb \xi^\mu)\ge 0, 
~ \forall i=1,\dots,N,
\end{equation}
in such a way that, if at time $t$ the system is prepared  or occurs to be precisely in that state, it will be there trapped for all $t'\ge t$. Of course, the fulfilment of \eqref{eq:stability} implies that $\bb J$ must be a suitable functional of $\bb \xi$. 
More generally, one is interested in assessing the convergence to a retrieval configuration even when the input quality is relatively low, namely, even when the initial configuration is relatively far (in Hamming sense) from the target pattern. In this case, one asks for a stronger condition, that is
\begin{equation}\label{eq:strong}
	\xi^\mu_i  \varphi_i(\bb \xi^\mu) \geq \kappa >0, ~ \forall i=1,\dots,N,
\end{equation}
which means that, although $\bb \sigma(t)$ displays some discrepancies with respect to $\bb \xi^{\mu}$, the dynamics \eqref{eq:zeronoise} is still ensured to converge to $\bb \xi^{\mu}$. In this inequality (corresponding to the basic requirement in Gardner's theory \cite{Gardner-JPA1988,GardnerDerrida-JPA1989,Personnaz-1985}), $\kappa$ controls the width of the attraction basins, \red{that is, if $\bb\sigma$ belongs to a Hamming ball $\mathcal B(\bb\xi^\mu, R(\kappa))$ centered in $\bb\xi^\mu$ with radius $R(\kappa)$, the network response will be $f(\bb\sigma) = \bb\xi^\mu$, with $f$ being the transfer function $f(\bb\sigma)=\lim_{n\to\infty} \mathcal T^{n}(\bb\sigma)$ and $\mathcal T(\bb\sigma)=\text{sgn}(\varphi_i (\bb\sigma))$ the 1-step dynamics}. Increasing $\kappa$, \red{the stability criterion will be satisfied in a ball with larger and larger radius $R(\kappa)$ surrounding} the patterns, but this goes at cost of a smaller amount of storable information vectors, resulting in a lower critical storage capacity.
{In particular, for symmetric networks ($\boldsymbol J = \boldsymbol J^T$) the largest number of patterns that can be retrieved is $N$ \cite{Gardner-JPA1988}.} 
In order to satisfy 
the inequality constraint \eqref{eq:strong}, we can impose a (stronger) equality condition requiring that, given $\gamma \geq \kappa$, 
\begin{equation}\label{eq:stronger}
	\xi^\mu _i \sum_{j\neq i } J_{ij}\xi^\mu_j=\gamma, ~ 
 \forall i=1,\dots,N,
\end{equation}
with $\gamma$ being the same for all the patterns. The latter point could appear a rather strong assumption, but -- at least in the random theory, where patterns are all equivalent, as their entries are i.i.d. -- it is reasonable. The requirement \eqref{eq:stronger} has important technical consequences. First, if the patterns are Boolean, it can be rewritten in a more transparent form as $\sum_{j\neq i} J_{ij}\xi^\mu_j = \gamma \xi^\mu_i$, which is nothing but the Personnaz {\it et al's} stability criterion \cite{Personnaz-1985,Kanter-Sompolinsky-1987}. Further, we can remove the constraint on the absence of self-interactions and allow for $j=i$ in the last sum\footnote{This has quantitative effects on the estimate of the critical storage capacity, that can be neglected here as the focus is rather on the generalization capabilities in an unsupervised scenario, see e.g. \cite{Kanter-Sompolinsky-1987}.},
thus recasting the previous expression as an eigenvalue problem as $\bb J\cdot \bb\xi^\mu = \gamma \bb\xi^\mu$; this means that the coupling matrix is designed so that the patterns are eigenvectors with degenerate eigenvalue $\gamma$.
\par\medskip
In the general case, we can add an external field by replacing the local internal field in \eqref{eq:strong} with the total field, i.e., $\varphi_i (\bb\sigma)\to \sum_{j } J_{ij}  \sigma_j +h_i$. The previous arguments still hold and, for $\mu=1,\dots,P$, our problem takes the form
\begin{equation}\label{eq:rigid_problem}
	\begin{cases}
		\bb J \cdot \bb \xi^{\mu} + \bb h =\gamma \bb \xi^{\mu}\\
		\bb J=  \bb J^T\
	\end{cases}.
\end{equation}
{
Therefore, in this context, training a network implies finding an arrangement for $\bb J$ and $\bb h$, such that \eqref{eq:rigid_problem} hold and this can be recast into the minimization of a Mean-Squared Error (MSE) of the form $(\bb J \cdot \bb \xi^\mu + \bb h -\gamma \bb \xi^\mu)^2$, in such a way that}
we can set up the minimization procedure for a loss function reading as\footnote{The case without self-interactions can be recovered by adding a contribution $\sum_i \theta _i J_{ii}$ in the loss functions, where $\theta_i$ are Lagrange multipliers ensuring that the diagonal entries are set to zero. 
}
\begin{equation}\label{eq:loss_1}
	\mathcal L_{\bb\xi} (\bb J,\bb h)=\frac1{2P} \sum_{i,\mu} \Big(\sum_j J_{ij} \xi^\mu_j +h_i -\gamma \xi^\mu_i \Big)^2+ \frac1{2P} \sum_{j,\mu}\Big (\sum_i J_{ij} \xi^\mu_i +h_j -\gamma \xi^\mu_j \Big)^2+\epsilon_J \sum_{i,j} J_{ij}^2+\epsilon_h \sum_{i} h_{i}^2 ,
\end{equation}
with $\epsilon_J,\epsilon_h \in[0,+\infty]$. The second term in the r.h.s. is obtained starting from the first one and reverting the roles of $i$ and $j$; these two contributions account for the stability criterion and the symmetry constraint\footnote{{We recall that the symmetry constraint for the coupling matrix $\boldsymbol J$ is introduced for consistency with the statistical mechanics picture and to ensure that the sequential dynamics \eqref{eq:zeronoise} exhibits fixed points.}} in a non-rigid way. The successive two contributions are $L_2$-regularization terms for, respectively, $\bb J$ and $\bb h$, that protect the norms of these parameters from divergence during training as they are confined by the quadratic potentials. 
\par\medskip
The explicit form of the solution of the constrained system \eqref{eq:rigid_problem} can be achieved via gradient descent method $\bb{\dot{J}}= -\nabla_{\bb J} \mathcal L_{\bb\xi}$ and $\dot {\bb h}= -\nabla_{\bb h} \mathcal L_{\bb\xi}$, which yields
\begin{eqnarray}
-	\bb{\dot{J}}&=& 	\bb J \bb \Omega +\bb \Omega \bb J +   \bb h \bar{\bb \xi}^T+  \bar {\bb\xi}\bb h ^T-2\gamma \bb \Omega +2\epsilon_J \bb J 	,\label{eq:1_J}\\
-\frac1{2}	\dot{\bb h}&=& \frac {\bb J+\bb J^T}2 \bar{\bb \xi}+(1+\epsilon_h)\bb h -\gamma \bar{\bb\xi}\label{eq:1_h},
\end{eqnarray}
where
$$
\Omega_{ij} \equiv \frac 1P \sum_\mu \xi^\mu_i \xi^\mu_j,\quad \bar \xi_i \equiv  \frac1P\sum_\mu \xi^\mu_i,
$$
are, respectively, the Hebbian kernel\footnote{With respect to the standard notation here the prefactor $1/N$ is replaced with $1/P$.} and the mean value of the $i$-th entry over patterns. Notice that the matrices $\bb J$ and $\bb J^T$ satisfy the same differential equation \eqref{eq:1_J} since $\bb \Omega$ is symmetric, thus, if we consider initial conditions such that $\bb J(0)=\bb J(0)^T$, the symmetry is preserved for any $t> 0$ by uniqueness arguments. Hence, we can safely replace $\bb J$ to its symmetric part in \eqref{eq:1_h}. {The convergence condition of the discrete form of these dynamical equations is discussed in App. \ref{app:num}.}
\par\medskip
Before concluding this Section, it is worth highlighting that, as standard, the neural relaxation \eqref{eq:zeronoise} and the training dynamics \eqref{eq:1_J}-\eqref{eq:1_h} operate on different time scales, the former being much faster. Biologically, this is motivated by the fact that synaptic plasticity is much slower that neural activation and, in artificial neural networks, by the fact that {the machine is first trained and later used for task accomplishment.} For consistency with \eqref{eq:zeronoise} one should therefore write $\bb{\dot{J}} = \frac{d \bb J}{dt} \tau_{J}^{-1}$ and $\dot{\bb h} = \frac{d \bb h}{dt} \tau_{h}^{-1}$, with $\tau_{J, h} \gg 1$. However, under this adiabatic hypothesis, we can consider the network parameters $\bb J, \bb h$ as planted during the neural dynamics and separate the two dynamical problems in such a way that, when focusing on the synaptic evolution, $\tau_{J, h}$ can be set as unitary without ambiguity.

\section{Dreaming as regularization, regularization as early-stopping\label{sec:dreaming}}
The global minimum for $\mathcal L_{\bb\xi} (\bb J,\bb h)$ in \eqref{eq:loss_1} can be obtained by requiring the stability conditions $\boldsymbol {\dot{J}}= \bb 0$ and $\boldsymbol { \dot{h}}= \bb 0$ in Eqs.~\eqref{eq:1_J}-\eqref{eq:1_h} that give
\begin{eqnarray}
	\bb J \bb \Omega + \bb \Omega \bb J+ \bb h \bar{\bb \xi}^T+ \bb {\bar\xi}\bb h ^T-2\gamma \bb \Omega +2\epsilon_J \bb J &=&0,\\
	 \bb J \bar{\bb \xi}+(1+\epsilon_h)\bb h - \gamma \bar{\bb\xi}&=&0,
\end{eqnarray}
whose solution reads as
\begin{eqnarray}\label{eq:solution}
	\bb h &=& \frac1{1+\epsilon_h}(\gamma \bb 1-\bb J)\bar {\bb \xi}\label{eq:h_sol},\\
	\bb J &=& \frac1P \hat {\bb \xi} \frac{\gamma}{\bb C +{\epsilon_{J}}\bb 1}\hat{\bb \xi}^T,\label{eq:J_sol}
\end{eqnarray}
where
$$
\hat{ \xi}^\mu_i \equiv   \xi^\mu_i -\Big(1-\sqrt{\frac {\epsilon_h}{1+\epsilon_h}}\Big)\bar { \xi}_i, \quad C_{\mu \nu} \equiv  \frac{1}{P} \sum_i \hat{\xi}_i^{\mu} \hat{\xi}_i^{\nu}.
$$

By inspecting Eq.~\eqref{eq:h_sol} one can see that the external field stems from the presence of biases in the input data, i.e. $\bar \xi_i \neq 0$, in such a way that re-centering the patterns by $\xi_i^{\mu} \to \xi_i^{\mu} - \bar{\xi}_i^{\mu}$ results in $\bb h = \bb 0$. Thus, as long as data are pre-processed in this way, external fields are not needed.
Further, by looking at Eq.~\eqref{eq:J_sol}, one can see that our solution recovers the interaction matrix of the ``Dreaming Hopfield model'' (DHM) \cite{FachechiNN, AABF-JStat2019,AlbertIEEE} 
\begin{equation} \label{eq:JD}
\bb J^{(D)} \equiv \frac{1}{P} \bb \xi \frac{t_d}{\bb I + \bb C t_d}\bb \xi^T,
\end{equation}
upon setting $\gamma= 1$ and identifying $t_d$ as the inverse of the hyperparameter $\epsilon_J$\footnote{\red{The minimization of the regularized-MSE as defined in \eqref{eq:loss_1} is a special setting of the usual setup of ridge regression theory \cite{horel1962application,hoerl1970ridge}, with the target response of the network being the multiplication of the input vectors $\bb\xi^\mu$ by the constant $\gamma$. Neglecting the bias vector, the ridge estimator minimizing the loss-function does coincide with the coupling matrix $\bb J^{(D)}$, where $t_d$ plays the role of Tichonov regularization parameter \cite{tikhonov1977solutions}. Ridge regression, together with their generalization to non-linear regression problems with kernel techniques \cite{vapnik1998statistical,scholkopf2002learning,vovk2013kernel}, is a central topic in statistical learning theory, focusing in particular on the role of the corresponding hyper parameter (see e.g. \cite{meanti2022efficient,alberti2021learning,wu2020optimal}) as well as the implicit regularization phenomena emerging in high-dimensional statistics \cite{hastie2022surprises,bartlett2020benign}.}}, that is, $t_d=\epsilon_{J}^{-1}$. 
More precisely, in the DHM, the kernel reads as $\bb {\tilde{J}}^{(D)} \equiv \frac{1}{N} \bb \xi \frac{t_d+1 }{\bb I + \bb C t_d}\bb \xi^T$ and it was obtained from the standard Hebbian kernel by iteratively applying an unlearning protocol based on an interplay of remotion and consolidation mechanisms inspired by those occurring during sleep in mammals' brain \cite{Hopfield-1983,crick1983function}; 
because of this analogy the time $t_d$, which measures the number of unlearning iterations, is referred to as ``dreaming time''.
{The critical storage of the DHM has been shown to increase monotonically with $t_d$, reaching, in the $t_d \to \infty$ limit, the theoretical upper bound known for symmetric networks and corresponding to a number of retrievable patterns equal to the number of neurons, i.e., $P=N$. Also, the DHM has been proved to outperform the standard Hopfield model as for generalization abilities} \cite{LaD}.
The difference between $\bb {J}^{(D)}$ and $\bb {\tilde{J}}^{(D)}$ just lays in a pre-fractor $P/N$ and in a shift $t_d \to t_d +1$ which, as long as they are finite and non-vanishing, yield only a quantitative correction.\footnote{In particular, the two models hold statistical mechanical equivalence as the partition function of the present model 
$Z_{\beta}(\boldsymbol{J}^{(D)}):=\sum_{\{\boldsymbol{\sigma}\}}\exp\left(-\beta\sum_{i,j}J^{(D)}_{ij}\sigma_{i}\sigma_{j}\right)$ can be turned into that of the original {DHM by rescaling} $\beta\to \frac{P}{N}  \frac{1+t_{d}}{t_{d}}\beta$, {where $\beta$ tunes the degree of stochasticity in the system, that is, it tunes the broadness of the distribution of the neural configurations or, in a physical jargon, it plays as the inverse temperature.}}
\newline
Moreover, for $t_d \to \infty$ (or, equivalently, for $\epsilon_J \to 0$), we recover Kohonen's projector matrix \cite{Kohonen-1984} 
$\bb J^{(P)} \equiv \frac{1}{P} \bb \xi \bb C^{-1} \bb \xi^T.$

\begin{figure}[tb]
	\centering
	\includegraphics[width=\textwidth]{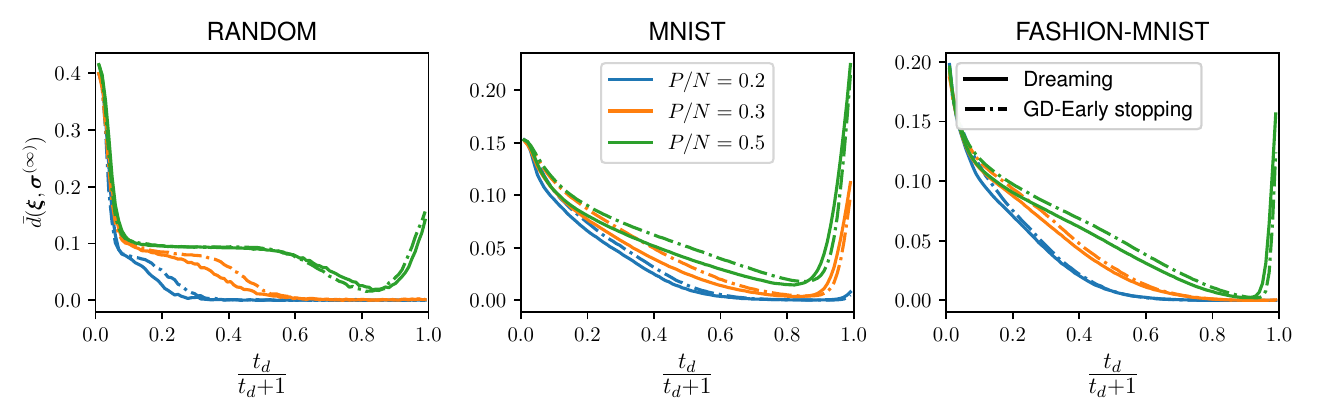}
	\caption{{\bfseries Retrieval performance of dreaming kernel {\it versus}  early-stopping.} The three panels show a comparison between the fully-trained solution \eqref{eq:J_sol} with $\epsilon_J \neq 0$ and the solution of the early-stopped training procedure with $\epsilon_J=0$; for the latter the final training time is chosen according to \eqref{eq:tstar}. In the leftmost panel, the dataset $\bb \xi$ is made of {$P$} Rademacher vectors that naturally display zero mean, while in the central and in the rightmost  panels {the dataset $\bb \xi$ is made of $P$ items randomly drawn from, respectively, the MNIST and the Fashion-MNIST datasets, and these vectors were pre-processed by Otsu method \cite{otsu} to make them binary. The items in these datasets were used to build up the interaction matrices $\bb J^{(D)}$ and $\bb J(t^*)$}. For the random dataset $N=200$ and $\gamma=1$, whereas for the MNIST and Fashion-MNIST $N=784$ and $\gamma=1$, also, {different values of the ratio $P/N$ are considered as reported in the common} legend. The performance of the system is measured in terms of the normalized Hamming distance $ d(\boldsymbol \xi^{\mu}, \bb \sigma^{(\infty)})$ between the target pattern $\boldsymbol \xi^{\mu}$ and the final configuration $\boldsymbol \sigma^{(\infty)}$, obtained by initializing the system in a corrupted version of $\boldsymbol \xi^{\mu}$ (obtained by flipping the pattern entries with probability $q=0.1$) and iterating \eqref{eq:zeronoise} up to convergence. {By averaging over all the $P$ patterns we obtain $\bar{d}(\bb \xi, \bb \sigma^{(\infty)}) = \frac{1}{P}\sum_{\mu} d(\boldsymbol \xi^{\mu}, \bb \sigma^{(\infty)})$, which is} plotted versus the dreaming time. {We refer to App.~\ref{metodi} for further details on numerics}. }\label{fig:Jd_vs_Jes}
\end{figure}

\begin{figure}[tb]
	\centering
	\includegraphics[width=0.6\textwidth]{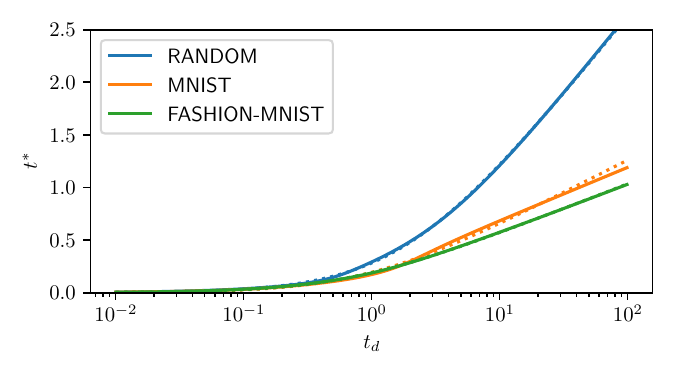}
	\caption{{\bfseries Stopping time as a function of dreaming time.} {The plot shows the early-stopping time $t^*$ as a function of the dreaming time $t_d$, obtained by a numerical estimate (solid line) from Eq.~\eqref{eq:tstar} and by a fit (dashed line) based on the functional relation} $t^*(t_d)=a \log (1+b\, t_d)$, suggested by the analytical findings presented in App.~\ref{sec:approxtstar}. The network parameters for the three cases are {$\gamma=1$, $N=784$, and $P/N=0.2$. The couple of coefficients $(a,b)$ estimated via  linear least squares are $(a=0.66, b=0.54), (a=0.11, b=3.25), (a=0.19, b=1.67)$ for the random, MNIST and Fashion-MNIST datasets, respectively.} \label{fig:ts_vs_td}}
\end{figure}

\medskip
Let us now move forward and notice that the solution $\bb J^{(D)}$ obtained by a fully-trained ($t \to \infty$) $L_2$-regularized ($\epsilon_J \neq 0$) process can be related to the solution of an unregularized ($\epsilon_J =0$) process which is run up to a finite time $t^*$; as we will see, this relation allows us to map the dreaming time $t_d$ into the training time and therefore interpret the dreaming mechanism as a training. In order to establish this relation, we resume the dynamical problem with recentered patterns, in such a way that the inferred field is vanishing\footnote{
Equivalently, we can choose to work without rescaling the patterns, thus also including the external fields. In this case, we can simplify the analysis of the dynamical problem by requiring that $\tau_J \bb {\dot J}= -\nabla _{\bb J}\mathcal L$ and $\tau_h \dot{\bb h} = - \nabla_{\bb h} \mathcal L$, and consider the case $\tau_h \ll \tau_J$. Under this assumption, the variation of the external fields is much faster than the typical evolution of the coupling matrix, so -- when dealing with the temporal behavior of the latter -- the fields do relax instantaneously towards their fixed point at fixed $\bb J(t)$:
$$
\bb h_\infty [\bb J (t)]= \frac1{1+\epsilon_h}(\gamma \bb 1-\bb J(t))\bar {\bb \xi}.
$$
As a consequence, the synaptic dynamics is described by the equation
\begin{equation*}
-\bb{\dot{J}}=\bb J(\hat {\bb \Omega}+\epsilon_{J}\bb 1)+(\hat{\bb \Omega}+\epsilon_{J}\bb 1)\bb J-2\gamma \hat{\bb\Omega},
\end{equation*}
with $\hat { \Omega}_{ij}=  \Omega_{ij} -  M_{ij} = P^{-1}\sum_\mu \hat \xi^\mu_i \hat \xi^\mu _j$ and $M_{ij}=(1+\epsilon_h )^{-1} \bar \xi_i \bar \xi_j$.
{When dealing with structured patterns, we will preserve the inferred field in order not to alter the graphical appearance of the data; this is of course completely irrelevant when dealing with a zero-mean data set.}}
and Eq.~\eqref{eq:1_J} simply reads as
\begin{equation}
-\bb{\dot{J}}=\bb J(\bb \Omega+\epsilon_{J}\bb 1)+(\bb \Omega+\epsilon_{J}\bb 1)\bb J-2\gamma \bb\Omega.\label{evoluJ}
\end{equation}
This can be recast in the basis of the eigenvectors of $\bb \Omega$ (denoting with $a,b=1,\dots,N$ the corresponding indices) as
\begin{equation}
-\dot{J}_{ab}=\left(\lambda_{a}+\lambda_{b}\right)J_{ab}+2\epsilon_{J}J_{ab}-2\gamma\lambda_{a}\delta_{ab},\label{evol}
\end{equation}
where $J_{ab}$ is the element of $\bb J$ in the current basis and $\sigma(\bb \Omega)=\{\lambda_{a}\}_{a=1}^{N}$ is the ${\bb \Omega}$ spectrum. By solving Eq.~\eqref{evol} we find that the non-diagonal terms asymptotically
go to zero as $J_{ab}\sim \exp[-t (\lambda_a +\lambda_b+2\epsilon_J)]$ for any initial condition, in such a way that, at the equilibrium point, the coupling matrix is diagonal. {Here, we choose to prepare the system in a configuration where no information is stored, i.e., a \emph{tabula rasa} setting $\bb J(t=0)=0$, in this way the off-diagonal entries remain stuck at zero at any time $t$. Remarkably, the diagonal structure of $\bb J(t)$, when expressed in the basis of the eigenvectors of $\bb \Omega$, implies that the two matrices share the same eigenvectors. As for the entries on the principal diagonal, given the above-mentioned initial condition,} the solution of the associated differential equation is
\begin{equation}\label{intothat}
J_{aa}(t)= \frac{\gamma\lambda_{a}}{\epsilon_{J}+\lambda_{a}}\Big\{1-\exp\big[-2t(\lambda_{a}+\epsilon_{J})\big]\Big\}.
\end{equation}
As expected, in the limit of large training-time, we recover $\bb J^{(D)}$, that is 
\begin{equation}
J_{aa}(t)
\underset{t\to\infty}=\frac{\gamma\lambda_{a}}{\lambda_{a}+\epsilon_{J}} =  J^{(D)}_{aa},\label{Jaa}
\end{equation}
while, expanding at small $t$, we get \begin{equation} J_{aa}(t)=\frac{\gamma\lambda_{a}}{\epsilon_{J}+\lambda_{a}}\Big \{1-\exp\big[-2t(\lambda_{a}+\epsilon_{J})\big]\Big\}\underset{t\ll1}{\approx}2t\gamma\lambda_{a}\end{equation} which corresponds to $\boldsymbol{J}\approx 2t\gamma\boldsymbol{\Omega}$: this means that, 
despite the blank initial condition, at the very start of the training, the kernel $\bb J$ {is close to} an Hebbian structure. 
\newline 
On the other hand, {by setting $\epsilon_J=0$ in Eq. \eqref{intothat}, we find that the diagonal terms evolve as}
\begin{equation}
J_{aa}(t)=\gamma[ 1-\exp({-2 \lambda_a  t})].\label{Jaa0}
\end{equation}
Now, we compare the two explicit forms of the coupling matrix, i.e. the regularized one at $t\to\infty$ \eqref{Jaa} and the $t$-dependent one with $\epsilon_J=0$ \eqref{Jaa0}, and search for the {characteristic} time $t^*$ at which the latter is as close as possible to the former. To do this, let us consider the quantity
$$
\delta(\lambda,t,\epsilon_J)= \gamma^2 \Big[\frac{\lambda}{\lambda+\epsilon_J}-1+\exp(-2t \lambda)\Big]^2,
$$
measuring the squared {difference} between the components in the two realizations at fixed eigenvalue $\lambda$. Then, we take the average over the ${\bb \Omega}$ spectrum, i.e.
$$
\bar{\delta} (t,\epsilon_J)= \int d\lambda\, \delta(\lambda,t,\epsilon_J)\rho_E (\lambda),
$$
where $\rho_E (\lambda)= \frac1N \sum_{\lambda_a \in \sigma({\bb \Omega})} \delta(\lambda-\lambda_a)$ is the empirical spectral distribution of ${\bb\Omega}$. Notice that $\bar{\delta}$ is nothing but the {squared} Frobenius distance between the Dreaming kernel and the unregularized coupling matrix. This quantity is minimized for the following  
\begin{equation}\label{eq:tstar}
t^* (\epsilon_J)= \underset{t}{\text{argmin}}\ \bar{\delta} (t,\epsilon_J).
\end{equation}
This relation provides an expression for the time $t^*$ at which the unregularized gradient descent over $\mathcal L_{\boldsymbol \xi}(\boldsymbol J, \boldsymbol h)$ should be interrupted if we want a coupling matrix close to $\bb J^{(D)}$ corresponding to the fully-relaxed, regularized gradient-descent. 
The equivalence between the two scenarios is validated for synthetic, MNIST \cite{mnist} and Fashion-MNIST \cite{fashion} datasets as reported in Fig. \ref{fig:Jd_vs_Jes}. 
\newline
The functional relation between $t_d = \epsilon_J^{-1}$ and $t^*$ highlighted in Eq.~\eqref{eq:tstar} is depicted in Fig.~\ref{fig:ts_vs_td}. {The logarithmic behavior is justified analytically in App.~\ref{sec:approxtstar}, where, by expanding \eqref{eq:tstar} around the mean eigenvalue, we obtain a first-order approximation of the early-stopping time $t^* (\epsilon_J)$ which depends on $t_d$ and on the trace of $\bb \Omega$.}
 
\medskip

{Another way to see the equivalence between regularization and early-stop is the following. Starting from Eq.~\eqref{intothat}, we notice that $\epsilon_J$ provides a natural scale as}
$$
J^{(D)}_{aa} \approx
\begin{cases}
 \gamma \Big(1-\frac{\epsilon_J}{\lambda_a}\Big)\quad \,\,\,\text{ if } \lambda_a \gg \epsilon_J\\
\frac{\gamma \lambda_a}{\epsilon_J}  \quad\quad\quad\quad \,\,\,\text{ if } \lambda_a \ll \epsilon_J
\end{cases}.
$$
{Therefore, in the regularization approach, the parameter $\epsilon_J$ prevents the saturation of all the diagonal entries of $\bb J^{(D)}$ to the value $\gamma$ (corresponding to Kohonen's projector $\bb J^{(P)}$\cite{Kohonen-1972,Kohonen-1984,Personnaz-1985,Kanter-Sompolinsky-1987}); in fact, as long as $\epsilon_J >0$ (or $t_d$ finite), only the entries corresponding to the top eigenvalues of $\bb\Omega$ get close to this limiting value}, while the others remain close to the initial condition ({i.e., $J_{aa}=0$ for any $a=1,...,N$).} On the other hand, the early-stop dynamically accounts for such a filtering, since the time $t^*$ at which we stop the training is chosen so that {only a subset of the diagonal entries}
saturate to the fixed point $J_{aa}^{(D)}=\gamma$ of the dynamical system \eqref{evoluJ}, while all the others are not changed in a substantial way w.r.t. the initial condition; {in fact, as highlighted by \eqref{Jaa0}, the characteristic time for saturation is entry-dependent and given by $(2 \lambda_a)^{-1}$, thus entries corresponding to large eigenvalues of $\bb \Omega$ are faster.}

\medskip

What we presented so far does apply to a general set of Boolean vectors, as the coupling matrix $\bb J$ naturally arises as the fixed point of a gradient descent algorithm, {the only hypothesis that we made on the vectors $\bb \xi$, that we want to store as attractors for the neural dynamics, being that they are of the same length and that they share the same ``importance'' $\gamma$}. In particular, the dataset items could represent (noisy) realizations of some unknown ground-truth patterns to which we have no direct access.
In this context, regularization -- preventing the network parameters from acquiring large norms during learning -- also allows for a reduction of the model specialization on the training set. The relation $t_d=\epsilon_{J}^{-1}$ therefore suggests that overfitting issues may arise for too large $t_d$, as we are going to {discuss in more details} in Sec.~\ref{sec:overfitting}. Further, the dreaming time $t_d$ {can} be related, through $t_d = \epsilon_J^{-1}$ and Eq.~\eqref{eq:tstar}, to the stopping time $t^*$ in unregularized versions of the gradient descent algorithm. This relation is consistent with the previous remark since early-stop techniques are indeed designed for avoiding overfitting\footnote{This is executed by following the behaviour of training and validation losses, abruptly stopping the training procedure when the latter exhibits a general growing behavior, see e.g. \cite{Boes-PRE1998}.} and, again, recalling the monotonic relation between $t_d$ and $t^*$, we expect that overfitting issues may arise for too large $t_d$.
{In the next section we will make use of the framework outlined in this section and, specifically, of the optimal interaction matrix $\boldsymbol J^{(D)}$,} in order to address the generalization capabilities of such models or, conversely, the emergence of overfitting.

\section{Emergence of \red{generalization and} overfitting in Hopfield-like models} \label{sec:overfitting}

\subsection{{A synthetic dataset}} \label{sec:dataset}
The results derived in Secs. \ref{sec:stability} and \ref{sec:dreaming} were obtained without making specific assumptions on the {binary vectors $\{\bb \xi^{\mu}\}_{\mu=1}^P$, however, in order to go further in the analytical investigations, some additional hypothesis are in order. In fact,} in theoretical studies one usually assumes that pattern entries are extracted according to a prescribed probability distribution that allows working out a controllable theory. For instance, when dealing with the Hopfield model, a common choice is to take pattern entries as i.i.d. Rademacher random variables, and thus treat the patterns as ground truths to be reconstructed starting from a corrupted version of them. However, in practical applications, one has no direct access to the ground-truth patterns, but only to empirical realizations constituting the dataset from which we want to extract information. In a supervised scenario, one knows {\it a priori} how the different items making up the dataset are partitioned between classes, so that it is possible to define class archetypes (for example, taking the average of examples belonging to the same category) which are taken as representatives of the ground vectors, see e.g., \cite{AAKBA-EPL2023,EmergencySN,Fontanari-1990}. In the unsupervised case, this clearly cannot be done, and the simplest way to proceed is to include all the examples, homogeneously, in the treatment. This is the path that shall be pursued in this section and hereafter we detail this unsupervised setting by considering a synthetic dataset. 
\par\medskip
Let $\{ \bb \zeta^\mu \}_{\mu=1}^K \in \{-1, +1\}^{N \times K}$ be the ground patterns to which we have access only through empirical realizations referred to as $\bb\xi^{\mu,A}$ with $A=1,\dots,M$ for each $\mu$.  We assume that the examples in this training dataset are obtained from the ground patterns with a multiplicative noise, that is, $\bb \xi^{\mu,A}= \bb \chi^{\mu,A}\bb \zeta^{\mu}$ (with entry-wise multiplication), with
$$
\textrm{Prob}[\chi ^{\mu,A}_i =\pm1 ]= \frac{1\pm r}{2},
$$
where $r \in[0,1]$ is the parameter quantifying the {\it quality} of the example (i.e., it measures the correlation between the example and the corresponding ground-pattern). 
Our \emph{training} dataset is therefore constituted by the set $\mathcal S = \{\bb \xi^{\mu,A}\}_{\mu=1,...,K}^{A=1,...,M}$ and we distinguish two loads: $\alpha:=KM/N$, i.e., the ratio between the number of examples in the {training set} and the network size, and $\eta:=K/N \leq 1$, i.e., the ratio between the number of classes in the dataset (which is in principle unknown) and the network size.
{With this kind of available information, we want to train the system in order to make it able to generalize, namely to reconstruct the hidden ground-truths $\bb\zeta^\mu$, starting from an input data $\bb\sigma^{(0)}$ that is a corrupted representation of $\bb\zeta^\mu$. Since we have no direct access to the ground-truths, a direct error minimization procedure is not feasible in this case. However, we can include each single item in our loss function and take advantage of emergent phenomena in Hopfield-like models: as we will see, for a sufficiently large dataset, a plethora of spurious states appear and, depending on the control parameters of the system, these can favour the appearance of a generalization phase. In this scenario, regularization mechanism plays a key role, preventing the solution of the system to trivialize or overspecialize on the training set. In this unsupervised setting, the interaction matrix is therefore obtained by plugging in \eqref{eq:JD}, that is still solution of the gradient descent Eq.~\eqref{eq:rigid_problem}, the empirical realization of the Hebbian kernel with}  
entries $\Omega_{ij} = \frac1 P \sum_{\mu,A} \xi^{\mu,A}_i \xi^{\mu,A}_j$, and the {empirical} correlation matrix, whose size is $(KM) \times (KM)$ and whose entries are $C_{(\mu A),(\nu B)}= \frac{1}{KM} \sum_i \xi^{\mu,A}_i \xi^{\nu ,B}_i$.

\par\medskip
{To evaluate the performance of the network, we generate a test set $\tilde{\mathcal S} = \{\tilde{\bb \xi}^{\mu,A}\}_{\mu=1,...,K}^{A=1,...,M}$, sampled in the same way as the training set, and initialise the network with the configurations of the test set, say $\bb \sigma^{(0)} = \tilde{\bb {\xi}}^{\mu,A}$, the latter being, by construction, a noisy version of $\boldsymbol \zeta^\mu$ with quality $r$. Next, we check whether the network response is $f(\bb\sigma^{(0)})=\bb\zeta^\mu$,} an outcome that we interpret as a correct generalization; conversely, the retrieval of one of the training items, {say $f(\bb\sigma^{(0)})=\bb \xi^{\mu, A}$}, is interpreted as overfitting. In the following 
{subsection, we discuss the role of spurious states in the emergence of generalization and overfitting.}

\subsection{Spurious states of training data enable generalization\label{sec:spurious}}

 In the classical Hopfield setup, spurious states (i.e., configurations that are combinations of stored data) are known to impair the retrieval capabilities of the model and should be suitably treated in order to reduce their attractiveness, see e.g., \cite{christos1996investigation,Hopfield-1983,crick1983function,Plakhov-1995,DYD-1991,FachechiNN,AlbertIEEE,AABF-JStat2019}. In fact, the dreaming mechanisms mentioned in Sec.~\ref{sec:dreaming} are precisely aimed at this purpose and their implementation improves the retrieval capabilities of the network. 
On the other hand, when dealing with a dataset made of unlabelled examples, the situation is quite different, and spurious attractors are crucial for the emergence of generalization capabilities of the network, as we are going to discuss.

\begin{figure}[tb]
	\centering
	\includegraphics[width=0.55\textwidth]{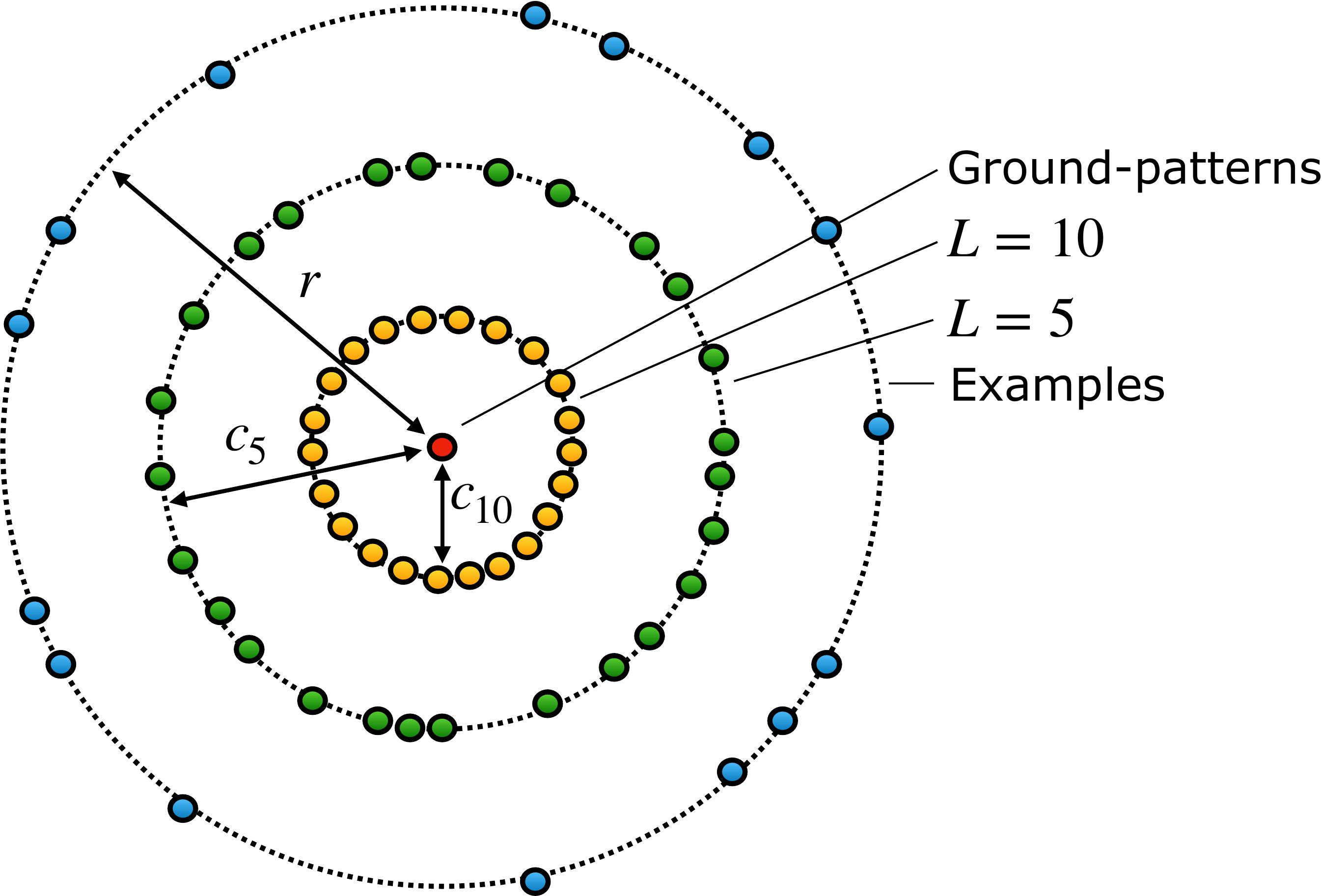}
	\caption{{\bfseries Schematic representation of training points, spurious combinations and ground-pattern.} The figure {sketches} the organization of attracting configurations within each class in the dataset. The class is represented by a ground-pattern $\bb\zeta$ (the red dot in the center), while the training points are located at distance $(1-r)/2$ from it (i.e. they have correlation $r$). Spurious combinations of training points are themselves attracting points, and their correlation $c_L(r)$ increases {with} the number $L$ of training points involved in the combination. For large enough $M$, the resulting landscape consists in many local minima very close to each other, so that they coalesce and form flat valleys around the ground pattern.}\label{fig:schema}
\end{figure}
Let us suppose that the training set is made of a large number $M$ of data for each class, and let us consider a spurious configuration given by a symmetric combination of $L$ examples pertaining to the same class, that is,
	\begin{equation}\label{eq:spurious}
		{\bb \xi}^\mu_L= \text{sgn} \Big( \sum_{l=1}^L {\bb \xi^{\mu,A_l}} \Big),
	\end{equation}
	being $A_1,\dots,A_L\in \{1,\dots,M\}$ the indices of the examples that we are mixing. 
	Denoting with $c_L(r):= {\boldsymbol \xi}_L^{\mu} \cdot \boldsymbol \zeta^{\mu} /N$ the correlation between ${\bb \xi}^\mu_{L}$ and the related ground truth, we
	notice that, as long as $L$ is relatively large, $ {\bb \xi}^\mu_{L}$ displays a correlation with the ground pattern $\boldsymbol \zeta^{\mu}$ that is larger than the correlation $r$ displayed by any training item, that is $r < c_L(r) \underset{L \gg 1}{\to} 1$, see App.~\ref{app: spurious} for more details.
\\
\red{Now, spurious configurations of the form \eqref{eq:spurious} in Hopfield-like models can be stable attractors, so that running the dynamics \eqref{eq:zeronoise} we could end in one of these minima and reach a relatively fair retrieval. Indeed, each of these configurations originates as a combination of attractors associated to stored vectors. In our scenario, increasing $M$, would result in an increasing number of intra-class spurious configurations which, as $L$ is increased, do exhibit larger and larger correlation with the corresponding ground-truths as sketched in Fig.~\ref{fig:schema}. For sufficiently large $M$, it is then reasonable to expect that minima of training examples and spurious configurations coalesce together, so that the resulting landscape consists in a wide minima centered in the ground-truth $\bb\zeta^\mu$, favoring the reconstruction of the hidden patterns.}

\begin{figure}[h!]
	\centering
 \includegraphics[width=\textwidth]{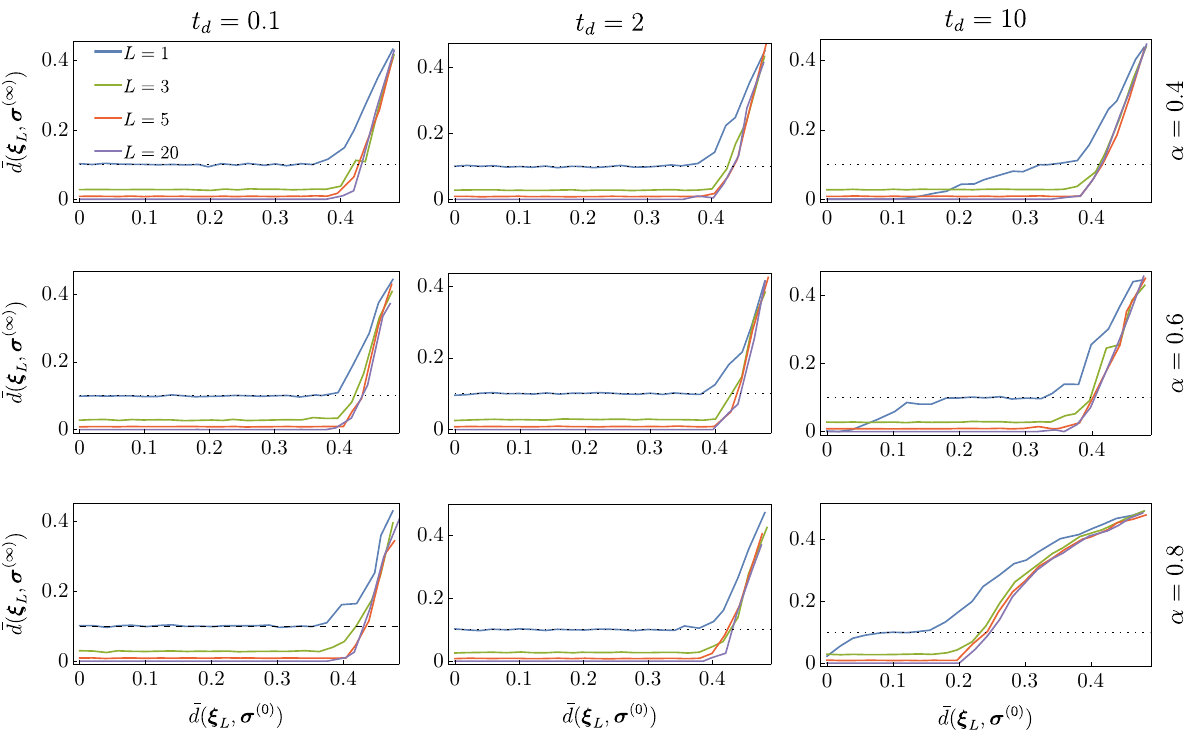}
	\caption{\red{{\bfseries Relaxation to fixed points from perturbed training examples and spurious states.} The plots show the retrieval capabilities of the model initialized in a configuration consisting in a perturbed version of the training examples (i.e., $L=1$) or in an intra-class spurious configuration $\bb \xi_L$ (as given by Eq. \eqref{eq:spurious}). The network size is fixed to $N=500$, the number of classes is $K=10$, and the quality of the dataset is $r=0.8$, while different values of dreaming time (from left to right $t_d=0.1,2,10$) and of load (from top to bottom $\alpha=0.4,0.6,0.8$, that is, $M=20,30,40$) are considered. The analysis is performed by taking a reference configurations ${\boldsymbol {\xi}}_L$ (with for $L=1, 3, 5, 20$, as explained by the legend) and applying a perturbation that consists in randomly flipping a fraction $q$ of the entries; preparing the system in this configuration $\bb \sigma^{(0)}$, we update the network  up to convergence towards the fixed point $\bb \sigma^{(\infty)}$. Then, we compare the average distances $\bar d(\bb\xi_L, \bb\sigma^{(0)})$ and $\bar d(\bb\xi_L, \bb\sigma^{(\infty)})$ between the reference configurations and, respectively, the initial and the final configurations. The dashed black lines correspond to the distance between the training examples used to build $\boldsymbol J^{(D)}$ and the associated ground-truths. The results are averaged over 50 different realizations of the dataset.
 }
 }\label{fig:attraction_basins}
\end{figure}

Given such a landscape, which is the role of $t_d$ (or, equivalently, of $\epsilon_J$)?
As recalled at the beginning of this subsection, in a Hopfield model where we store ground patterns, dreaming mechanisms reduce (and ultimately remove, if the load is not too high) the stability of spurious mixtures between independent patterns and this is obtained by shrinking {and lifting} the attraction basins associated to the patterns. Moving to an unsupervised setting, we realize that there are two kinds of mixture, according to whether they involve examples belonging to different classes or examples belonging to the same class; the former, just like in the Hopfield model, impair retrieval and should be removed, while the latter, as mentioned above, can be beneficial for generalization. Therefore, in this case, the dreaming mechanism should operate in removing only the first type of correlation. In fact, by increasing $t_d$ we are disentangling the minima corresponding to the stored patterns and this process affects progressively minima with larger and larger overlap. Inter-class correlation is typically smaller than intra-class correlation -- their extents being related to, respectively, $K/N$ and $r$ -- and relatively small values of $t_d$ can be sufficient to detach the attraction basins related to different ground-patterns. Yet, if we let the dreaming mechanism operate for too long a time, {the m\'elange of intra-class minima can be separated as well and they get fragmented in many energetic minima, each associated to a single example.}
As a consequence, we cannot retrieve spontaneously-formed archetypes, but only the single examples: the system is specialized over the training set, thus ending in an overfitting regime. 
{
This picture is corroborated by the numerical results reported in Fig. \ref{fig:attraction_basins}. Here, we focus on the behavior of the system when prepared in the neighborhood of a training example or of an intra-class spurious configuration. First, we notice that, for low enough $t_d$, even when the system is initialised in a configuration consisting in a mild perturbation of one of the stored examples ($L=1$), the neural dynamics will drive it towards a final configuration with the relative distance w.r.t. the reference example being $(1-r)/2$ (precisely the distance between the stored examples and the corresponding ground-truth). For spurious states, the situation is similar, with the relative distance between the final state and the reference mixture getting lower and lower as $L$ is increased (as the correlation $c_L(r)$ with the ground-truth increases monotonically with $L$). A moderately larger $t_d$ yields a larger attractivity of the ground patterns. This picture is consistent with our claim about the coalescence of the population of attraction basins in a wide minima centered at $\bb\zeta^\mu$. By further increasing $t_d$, the training points get attractive, with a basin width depending on the number of examples per class. 
This signals that, according to the setting, the dreaming mechanism can either enhance generalisation or favour overfitting.}

\section{Numerical experiments}\label{sec:numerics}
In this Section, we provide numerical evidence to our theoretical findings. 
{We first inspect the regions in the space of parameters $(\alpha, t_d, r)$ where the system equipped with the interaction matrix $\bb J^{(D)}$ can successfully generalize, that is, when tested with examples not included in the training set (but sharing with them the same underlying statistics), it is able to fairly reconstruct the ground pattern. Next, we corroborate this picture by applying a clustering algorithm to the network outputs and showing that, in the region of the parameter space where the system is expected to generalize (resp. overfit), the number of classes is nicely estimated (overestimated). Details on numerics are collected in App.~\ref{metodi}.}  

\begin{figure}[tb]
	\centering
	\includegraphics[width=1.0\textwidth]{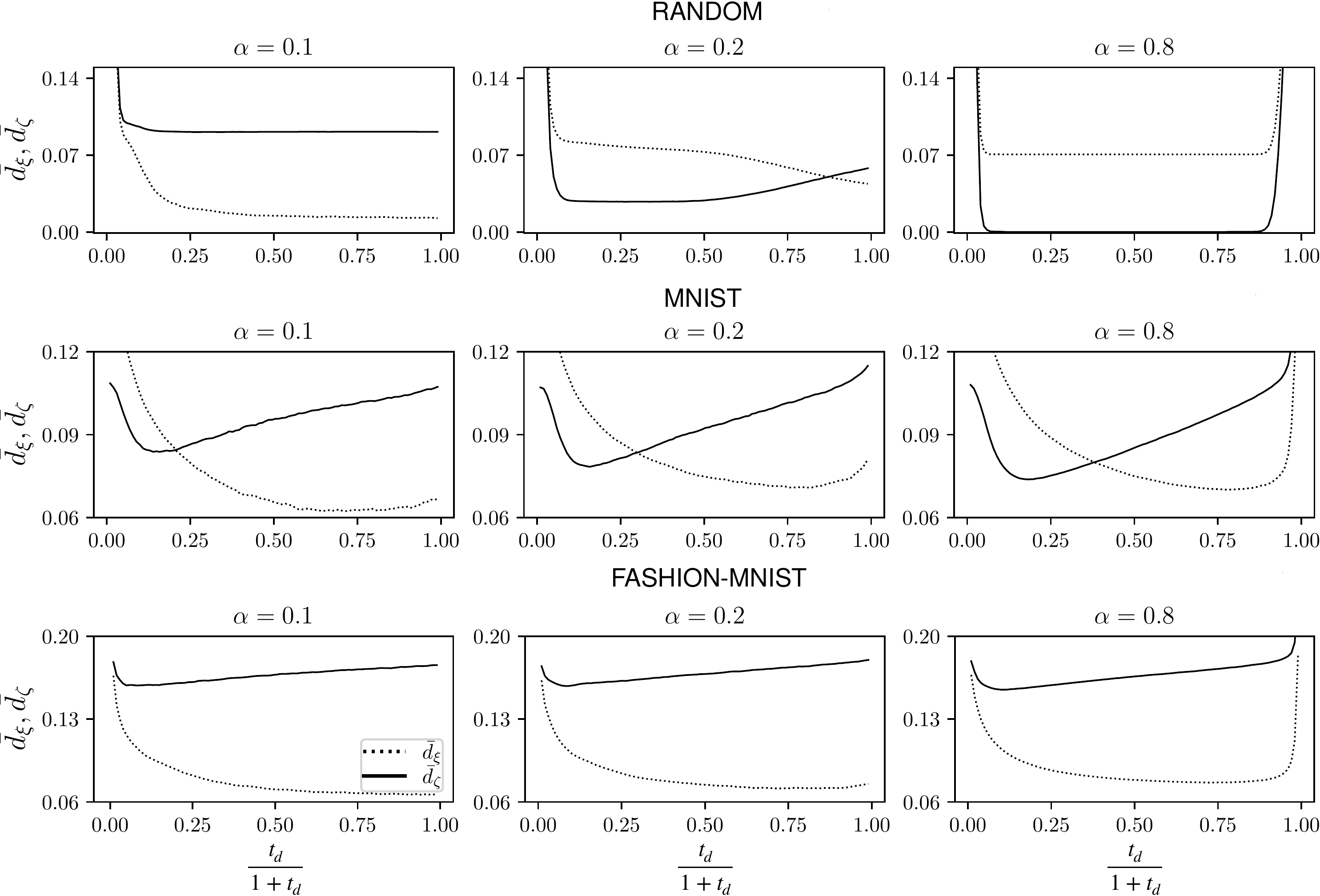}
	\caption{{\bfseries Retrieval on synthetic and structured datasets.} The retrieval performance is measured in terms of the normalized Hamming distance $d$ between the final configuration $\boldsymbol \sigma^{(\infty)}$ and {the nearest training example $\bb \xi$ (dotted curve, see Eq.~\eqref{eq:d_a}) and the nearest ground-truth $\bb \zeta$ (solid curve, see Eq.~\eqref{eq:d_b}); the results presented have been averaged over the $K\times M$ different initial configurations which constitute the test set (see App.~\ref{metodi} for further details). The network parameters for the random dataset are $N=200$, $K=10$ and $r=0.8$, whereas for the structured datasets they are $N=784, K=10$. For all the datasets, we reported results for {different choices of $\alpha=0.1$, $0.2$ and $0.8$, retaining $\eta=K/N$ fixed, therefore, recalling that $\alpha=K M/N$, we varied $\alpha$ by increasing $M$.}  } }\label{complessivo}
\end{figure}
\subsection{Generalization diagrams}
{In the first part of the numerical experiments we consider both structureless and structured datasets, to confirm and check the robustness of our theoretical results. The structureless datasets are built synthetically as follows: we initially generate a set of $K$ Rademacher ground patterns $\mathcal G = \{\bb \zeta^{\mu}\}_{\mu=1,...,K}$, whence we obtain a set of training examples $\mathcal S = \{ \bb \xi^{\mu,A}\}_{\mu=1,...,K}^{A=1,...,M}$ (characterized by a quality $r$ as specified in Sec.~\ref{sec:dataset}), which are used to build $\bb J^{(D)}$, according to Eq.~\eqref{eq:JD}. Next, we generate a test set $\tilde{\mathcal S} = \{ \tilde{\bb \xi}^{\mu,A}\}_{\mu=1,...,K}^{A=1,...,M}$, applying the same procedure used for the training set, that is, each item $\tilde{\bb \xi}^{\mu,A}$ exhibits a quality $r$ w.r.t. the related ground pattern $\bb \zeta^{\mu}$. For structured datasets, we consider the MNIST \cite{mnist} and the Fashion-MNIST \cite{fashion} benchmarks and define the ground patterns as the class averages, then, the training and the test sets are made of $M$ items, drawn from the whole datasets (overall made of, respectively, $60000$ and $10000$ instances), in such a way that the two sets have null intersection.}

{Whatever the dataset, we initialize the system in a configuration $\bb \sigma^{(0)}$ belonging to the test set, we run the dynamics \eqref{eq:zeronoise} and collect the final configuration $\bb \sigma^{(\infty)} = f(\bb \sigma^{(0)})$. In other words, $\bb \sigma^{(0)}$ and $\bb \sigma^{(\infty)}$ represent, respectively, the input and the output of the system. Next, we evaluate the following quantities:}
\begin{eqnarray} \label{eq:d_a}
d_{\xi} &=&  \min_{\bb \xi \in \mathcal S} d(\bb \xi,\bb \sigma^{(\infty)}), \\
 \label{eq:d_b}
d_{\zeta} &=&  \min_{\bb \zeta \in \mathcal G} d(\bb \zeta,\bb \sigma^{(\infty)}),
\end{eqnarray}
{where $d: \{-1,+1 \}^N \times \{-1,+1 \}^N \to [0,1]$ is the normalized Hamming distance, measuring the fraction of misaligned entries among the two configurations that are compared.
We stress that, despite not highlighted in Eqs.~\eqref{eq:d_a}-\eqref{eq:d_b}, these quantities depend on the initial configuration as $\bb \sigma^{(\infty)}$ does depend on $\bb  \sigma^{(0)}$.}
{These distances are then averaged over the sample $\tilde{\mathcal S}$, to get, respectively, $\bar d_{\xi}$ and $\bar d_{\zeta}$.}
\par\medskip
{In Fig.~\ref{complessivo}, we compare the behaviour of $\bar d_{\xi}$ and $\bar d_{\zeta}$ versus $t_d$, for different choices of $M$, while $K$ is fixed.
We find that, in any case, $\bar d_{\xi}$ and $\bar d_{\zeta}$ are monotonically decreasing with $t_d$ as long as $t_d$ is relatively small, next, their behavior depends on $M$. In particular, for the random dataset, when $M$ is small, we always have $\bar{d}_{\xi} < \bar{d}_{\zeta}$ that evidences poor generalization capabilities; when $M$ is larger we can leverage $t_d$ to enhance the generalization capabilities and get $\bar{d}_{\zeta} < \bar{d}_{\xi}$, yet when large $t_d$ is too large the curves cross; finally, when $M$ is large, at intermediate dreaming time, $\bar{d}_{\xi}$ and $\bar{d}_{\zeta}$ exhibit a plateau and at large values of $t_d$ they grow, the height of the plateau (respectively, $\approx (1-r)/2$ and $\approx 0$) suggests that there the final configuration is close to the ground pattern, while the final growth suggests a possible harmful effect of a large dreaming time. As for structured datasets, $\bar d_{\zeta}$ exhibits a minimum at intermediate values of $t_d$, corresponding to an optimal generalization performance and, again, for large dreaming times, $\bar{d}_{\zeta}$ grows. In particular, for the MNIST dataset, for relatively small (resp. large) values of  $t_d$, we get $\bar d_{\zeta} < \bar d_{\xi}$ (resp. $\bar d_{\zeta} > \bar d_{\xi}$), suggesting good (impoverished) generalization capabilities. Before proceeding, we also emphasize that, for all the datasets considered, when $t_d \gg 1$ and when $M$ is relatively large, both $\bar{d}_{\zeta}$ and $\bar{d}_{\xi}$ grow. This is due to the fact that the number of examples is relatively large to give rise to spurious attractors, but not large enough to make these attractors close to the ground truths; this point is further examined in the following. }

\begin{figure}[tb]
	\centering
 \includegraphics[width=0.7\textwidth]{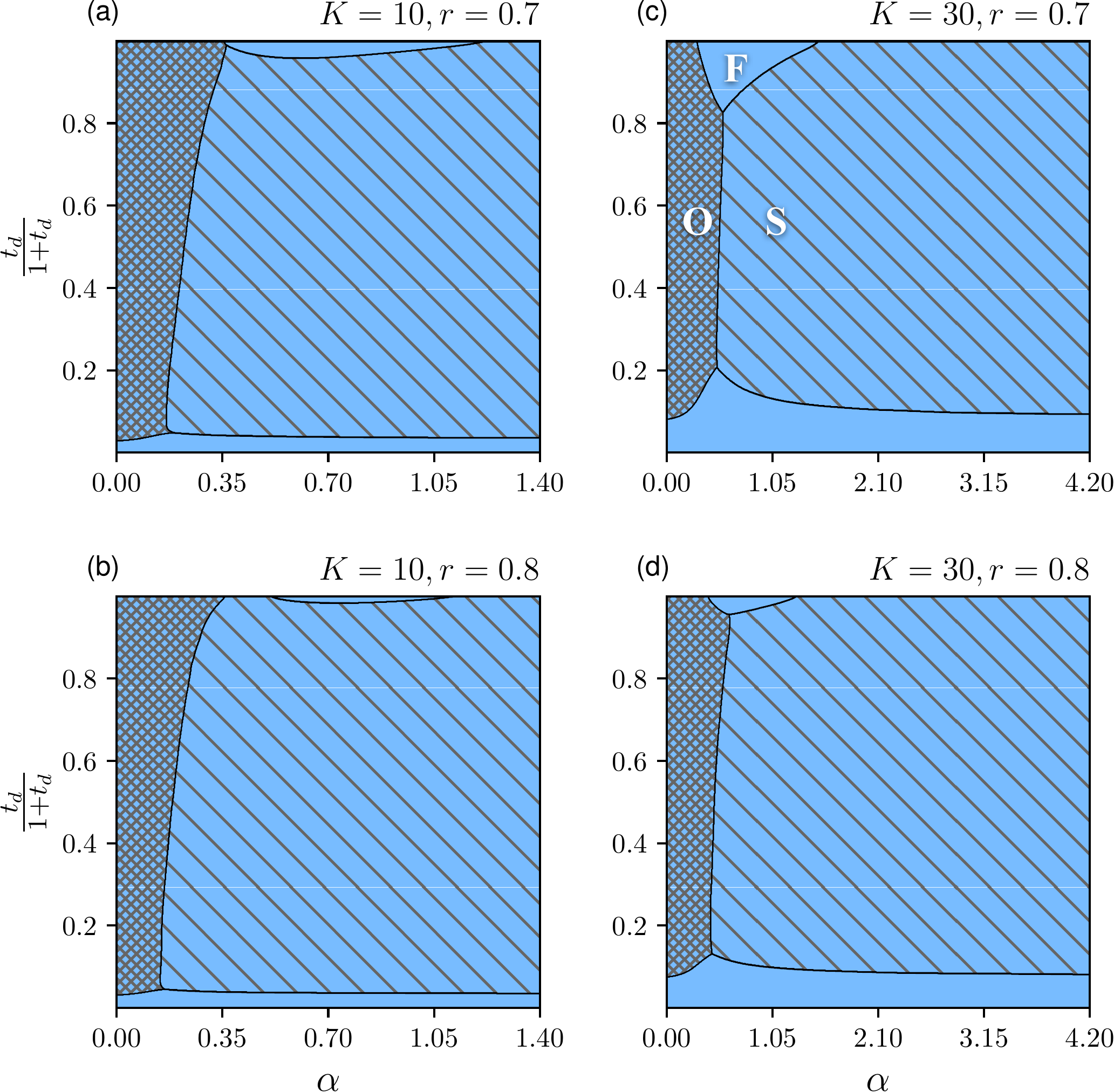}
	\caption{{\bfseries Generalization diagrams.} The four diagrams show the generalization outcomes of the neural network where the interaction matrix $\boldsymbol J^{(D)}$ is built on a sample of {random, synthetic} examples $\{ \boldsymbol \xi^{\mu,A}\}_{\mu=1,...,K}^{A=1,...,M}$ with $N=200$ {and $M$ tunable ($M=\alpha N/K$)}. In the $(\alpha,t_d)$ plane, for various values of $K$ and $r$, we outline three regions: success (S), overfitting (O), failure (F). {In any case the initial conditions $\boldsymbol \sigma^{(0)}$ are taken as perturbed versions of the ground-truths sampled with the same quality $r$ as the training examples.}\label{fig:diagrams}}
\end{figure}

{We now focus on the synthetic dataset and summarize the information processing capabilities of the system into ``phase diagrams''. To this aim, we distinguish between different outcomes as follows: }
\begin{itemize}
	\item {\bfseries Success}: this corresponds to {$\bar d_{\zeta} < \bar d_{\xi}$ \emph{and} $\bar d_{\zeta}<\frac{1-r}{2}$}. The first requirement ensures that the system relaxes in a configuration which is more correlated with the ground pattern than with the training points; the second condition, instead, guarantees that the dynamics ends up within the Hamming ball centered in the ground-truth with radius $(1-r)/2$ {and therefore that the system has moved closer to the ground-truth}.
	\item {\bfseries Overfitting}: this corresponds to {$\bar d_{\zeta} \geq \bar d_{\xi}$ \emph{and} $\bar d_{\xi}<\frac{1-r}{2}$. The first condition states that the final configuration is closer to one of the training points than to the ground; the second condition, guarantees that, this time, the dynamics ends up within the Hamming ball centered in the nearest training points with radius $(1-r)/2$ and therefore that the system has moved closer to a specific training item. }
	\item {\bfseries Failure}: otherwise. In this case the system is neither sufficiently close to a ground-pattern nor to an example.
\end{itemize}
For a given choice of $K$ and $r$, moving within the $(\alpha,t_d)$ plane, we thus depict the generalization diagrams for synthetic datasets. The results are reported in Fig.~\ref{fig:diagrams} and discussed hereafter. 

Starting from panel $a$ ($K=10$ and $r=0.7$), we find that, at very low $t_d$, the system is always in a failure regime. This is not surprising since, there, the interaction matrix is very close to the Hebbian prescription and, for this choice of the dataset parameters, interferences among examples are significant enough for the system to be likely to end up in {inter-class spurious} states, thus we expect that equilibrium configurations are non-retrieval states. Increasing $t_d$, such interferences are (possibly) removed. If the load parameter $\alpha$ is low (meaning that the number of examples is low), the system enters in an overfitting regime because the minima corresponding to examples {are sparse enough to} be easily disentangled. On the other hand, by increasing $\alpha$ (namely by increasing the number of examples per class), the attractors coalesce and {configurations corresponding to the ground patterns} become more and more attractive, in such a way that the system starts to well-generalize. {When $t_d \gg 1$, as $\alpha$ is increased, the transition from overfitting to success is no longer direct as, for intermediate values of $\alpha$, we can end up into spurious states that are still too sparse to ensure a sound generalization; this region can be shrunk by enhancing the dataset quality.}
In fact, by increasing the dataset quality as in panel $b$ ($K=10$ and $r=0.8$), the qualitative picture is the same, with just an expansion of the success region due to the fact that {intra-class} examples are now closer to each other. Next, we move to panels $c$ and $d$, where a larger $\eta$ (namely, a larger number of classes $K=30$, with fixed $N$) {benefits the failure } 
region, since in this case the clusters of minima associated to each class are closer
and 
clusters {of minima corresponding to different classes} now present non-trivial overlaps, in such a way that the relaxation of the system could end far away from the class for which the initial condition was generated.

\subsection{An analogy with a clustering algorithm}

{
In this section, we use another approach to check the emergence of overfitting and generalization regimes as the system parameters are tuned. 
The idea is to use an unsupervised clustering algorithm to partition the final configurations $\bb \sigma^{(\infty)}$ obtained by applying the dynamics \eqref{eq:zeronoise} to the test configurations; here unsupervised refers to the fact that the clustering algorithm is unaware of the number of effective clusters.}

{
We start the experiment by generating a random synthetic datasets $\mathcal G$ made of $K$ Rademacher ground patterns, whence we build a training set $\mathcal S$ and a test set $\tilde{\mathcal S}$, both characterized by a quality $r$ and a size $M$. We use the former to construct $\bb J^{(D)}$ and the latter to initialize the neural configuration. We collect the final configurations $\bb \sigma^{(\infty)}$ obtained by iterating the neural dynamics and we expect that, if the network correctly generalizes, the clustering algorithm applied to the final configurations will return an estimated number of clusters $\hat{K}$ which is (approximately) $K$ and each cluster contains a number of items which is (approximately) $M$. Conversely, if the network overfits, we expect that $\hat K > K$. In order to quantify the likelihood of these outcomes we introduce the accuracy $\frac{\hat{M}}{M\times K} \in [0,1]$, where $\hat{M}$ is the total number of examples correctly clustered by the algorithm.}
{The unsupervised clustering algorithm considered here is based on the Disjoint Set Union data structure \cite{disjoint}, which works as follows. Initially, it associates to each item $\bb \sigma^{(\infty)}$ a different class, in such a way that, at this stage, the number of estimated classes is $M\times K$. Next, we consider all the $\binom{M}{2}$ couples of configurations $\bb \sigma^{(\infty)}$ and check whether their normalized Hamming distance is smaller than a threshold $d^*(r)$ and, if so, the two items are merged in the same class. Once all the couples have been examined the algorithm stops. The threshold value is chosen equal to the minimum of the normalised Hamming distance between all pairs of examples belonging to the test set: 
$$
d^*(r) = \min_{\substack{(\bb \sigma_1, \bb \sigma_2) \in \tilde{\mathcal S}\times \tilde{\mathcal S}\\ \bb \sigma_1 \neq \bb \sigma_2}} d(\bb \sigma_1, \bb \sigma_2).
$$
The idea underlying this choice is that, after applying the dynamics to the test examples, if the network performs well, the examples belonging to the same class get closer to the common ground pattern, their distance is reduced and expected to be smaller than $d^*(r)$; on the other hand, for two examples belonging to different classes   the distance is expected not to vary significantly and remain larger than  $d^*(r)$.}

\begin{figure}[tb]
	\centering
 \includegraphics[width=1.0\textwidth]{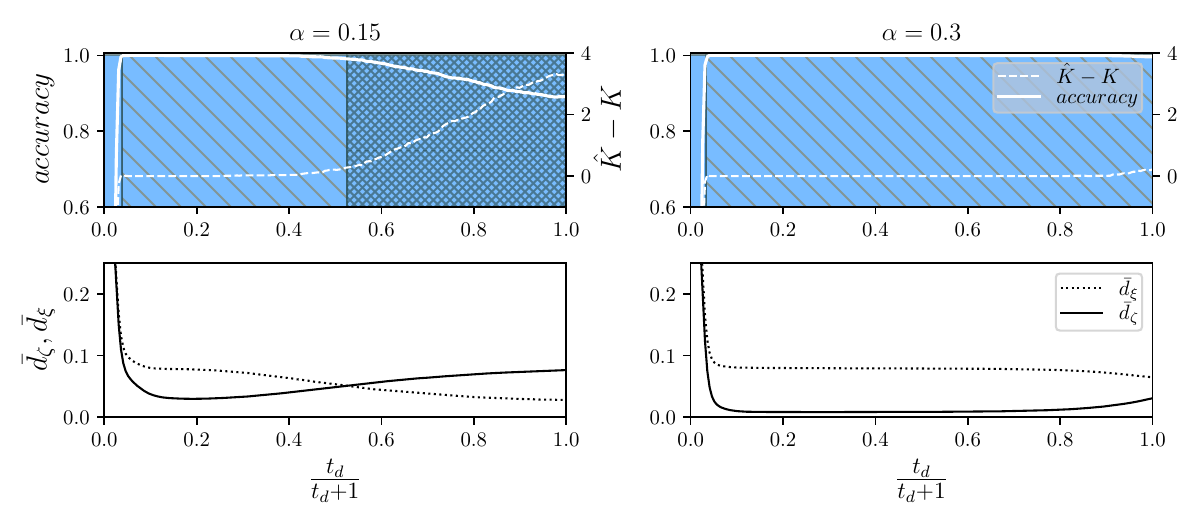}
	\caption{{\bfseries Clustering of the test set.} { Application of the clustering algorithm in case of a random synthetic dataset with parameters $K=10,r=0.8,N=200$ for two different loads values: $\alpha=0.15$ and $\alpha=0.3$ keeping fixed the number of ground-truths $K$. In the simulation with $\alpha=0.15$ the overfitting region occurs when  $\bar d_{\xi}$ is overcome by  $\bar d_\zeta$, in this region the clustering algorithm is no longer able to correctly cluster the examples and the number of clusters estimated starts to grow. In the simulation with $\alpha=0.3$, the number of examples per ground truth in the training set is double that of the previous simulation,  $\bar d_{\zeta}$ is always lower than that of the nearest training example $\bar d_{\xi}$ and the overfitting region is no longer present.}}\label{fig:diagrams_cluster}
\end{figure}

{
The results of this experiment, repeated for different loads and different dreaming times, are reported in Fig. \ref{fig:diagrams_cluster}.
The panels in the first row show the $accuracy$ (left axis) and the difference between $\hat K$ and the true number of classed $K$ (right axis) as a function of $\frac{t_d}{1+t_d}$, while the panels in the second row show the average distances $\bar{d}_{\zeta}$ and $\bar{d}_{\xi}$ between $\bb \sigma^{(\infty)}$ and, respectively, the ground truth and the nearest example, as a function of $\frac{t_d}{1+t_d}$. The colours in the background correspond to the different regimes of the network and we used the same colormap previously adopted in Fig.~\ref{fig:diagrams}, to highlight the consistency. In fact, in the failure region the accuracy is low and the number of clusters is underestimated; in the success region the accuracy is unitary and $\hat K = K$; in the overfitting region the accuracy is suboptimal and the number of cluster is overestimated. Further, these outcomes are nicely mirrored by the behavior of $d_{\zeta}$ and $d_{\xi}$: the transition from failure to success corresponds to an abrupt decrease of $d_{\zeta}$ that leaves $d_{\xi}$ behind; the transition between success to overfit corresponds to $d_{\xi}$ outpacing $d_{\zeta}$.}

\begin{figure}[tb] 
	\centering
 \includegraphics[width=0.75\textwidth]{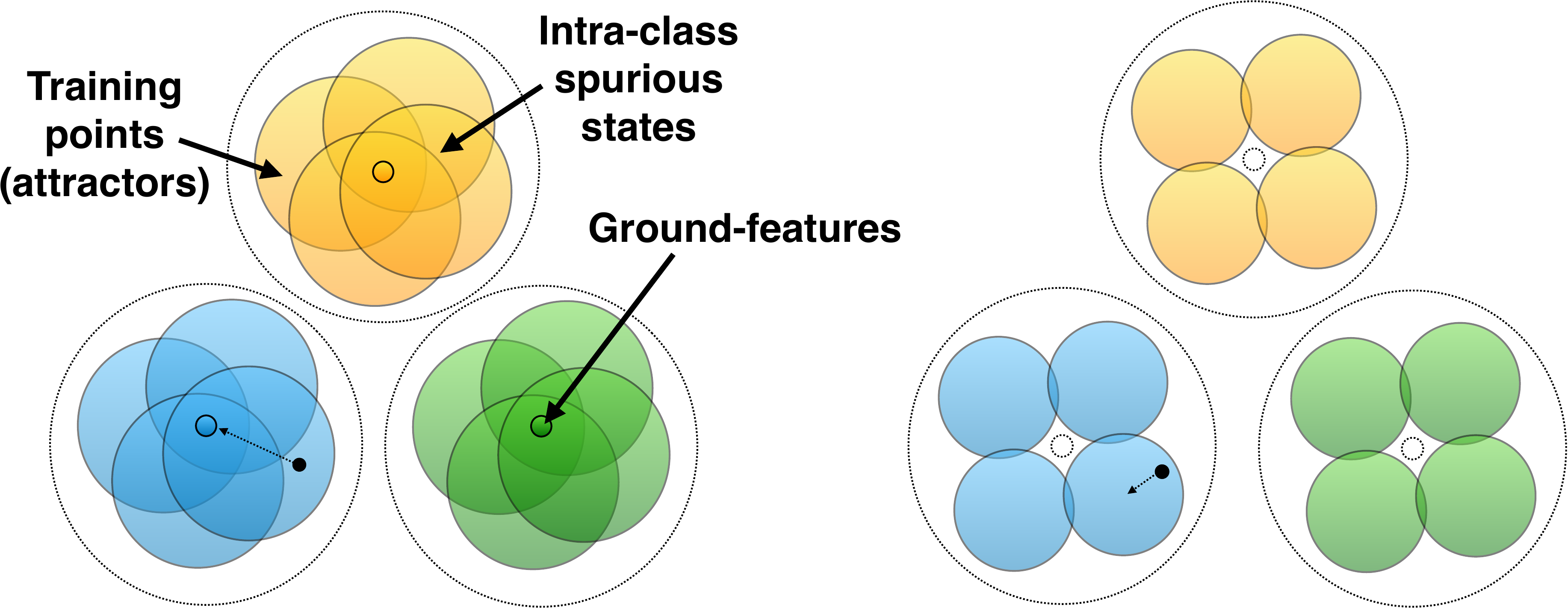}
\vspace{0.5cm}\\
 \includegraphics[width=0.8\textwidth]{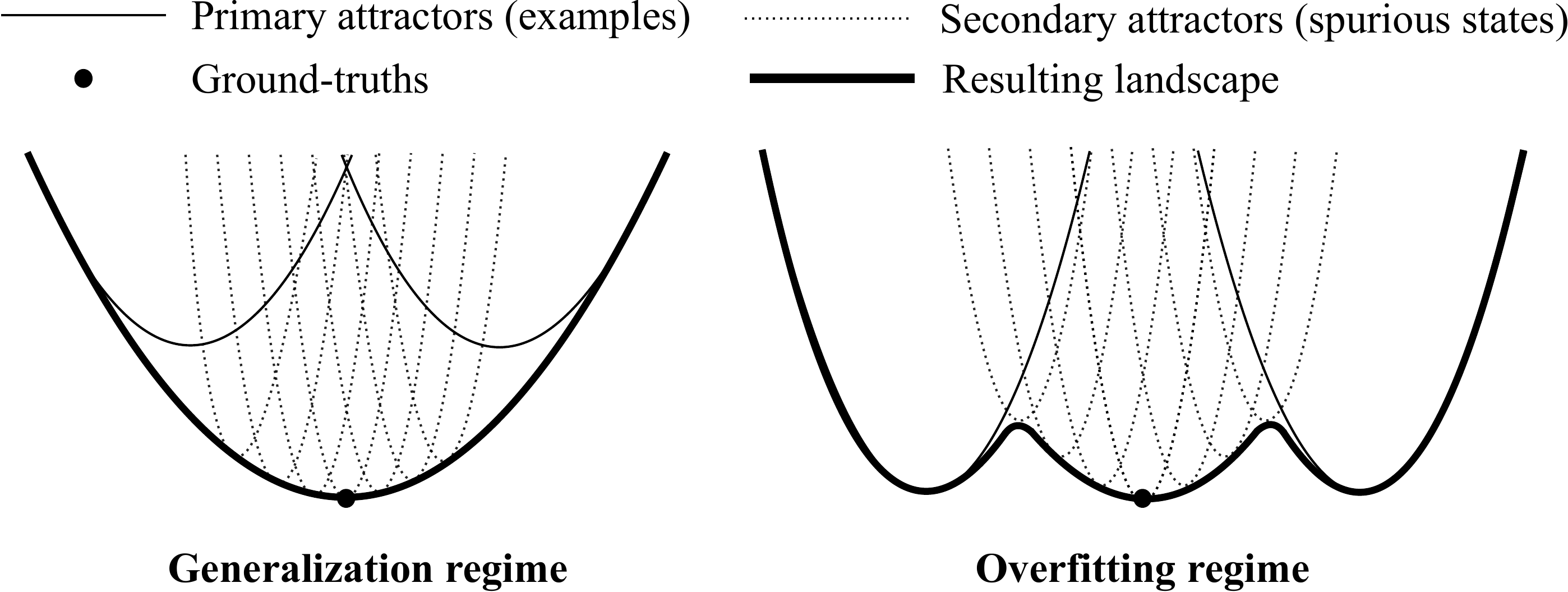}
	\caption{\textbf{Emergence of fixed points for a system trained without supervision}. {This schematic picture shows the evolution of the landscape generated by $\bb J^{(D)}$ as $t_d$ is varied. 
 When $t_d$ is relatively low (left), the m\'elange of minima has been partially disentangled: the weak inter-class interference is removed, while the minima corresponding to examples of the same class are still clusterized; when $t_d$ is relatively large (right) the interference among training examples has been completely shifted. This picture is compatible with our numerical results reported in Fig. \ref{fig:attraction_basins}.}
 \label{fig:sketch}}
\end{figure}

\section{{Conclusions}} \label{sec:conclusions}

{The main results obtained in this work are listed hereafter:
\begin{itemize}
\item We introduced a regularized loss function, whose minimum provides the interaction matrix $\bb J$ of an associative neural network; according to the dataset provided, the neural network equipped with the solution $\bb J$ is able to retrieve a set of stored ground patterns or to generalize starting from a corrupted version of unknown ground patterns.
\item We proved that the solution $\bb J$ of the loss function corresponds to a Hebbian-like kernel $\bb J^{(D)}$, known as dreaming Hebbian kernel and parametrized by the ``dreaming time'' $t_d$, as long as $t_d$ is identified with the inverse of the regularization parameter $\epsilon_J$.
\item In the absence of regularization ($\epsilon_J=0$) the solution $\bb J^{(D)}$ can be recovered by applying an early-stop strategy to the gradient descent over the loss function. This suggests that $t_d$ (or, equivalently, $\epsilon_J=0$) plays a role in preventing overspecialization on the training set.  
\item
Focusing on the case of a training set made of corrupted versions of some unknown ground patterns, we found robust numerical evidence that relatively large values of $t_d$ and relatively sparse training sets can yield overfitting. 
\item The emergence of overfitting is related to the structure of the Lyapunov function associated to the neural dynamics and this picture allowed us to speculate on optimal settings for the loss hyperparameters and/or for the training time. This picture is sketched in Fig.~\ref{fig:sketch}.
\end{itemize}}

To conclude, our results highlight the relevant mechanism allowing for the emergence of generalization capabilities of Hopfield-like networks: this is identified as the coalescence of attractors associated to training points giving rise to wide minima around the underlying ground truths (which is, indeed, a crucial ingredient for general models exhibiting robust generalization properties, see for example \cite{zecchina1,zecchina2} and references therein). In this scenario we give a comprehensive characterization of generalization and overfitting for synthetic random datasets. Developments of the present work would require the extensions to structured data, as well as a statistical mechanics characterization of relevant collective phenomena.

\section{Acknowledgments}
E.A. and A.F. acknowledge financial support from PNRR MUR project PE0000013-FAIR and from Sapienza University of Rome (RM120172B8066CB0, AR2221815D7192C1, AR1221815EA97525). \\ 
F.A. and A.F. have been fully supported by PNRR MUR project PE0000013-FAIR.\\
E.A., M.A., F.A., A.F. acknowledge the stimulating research environment provided by the Alan Turing Institute’s Theory and Methods Challenge Fortnights event “Physics-informed Machine Learning”.
\appendix

\section{Convergence of the gradient descent procedure \label{app:num}}
The discretization of the dynamical equation  \eqref{evoluJ} reads as
\begin{equation}\label{app:eqJ}
	 \bb J(n+1)= \bb J(n)-\epsilon\big[	\bb J(n)({\bb \Omega}+\epsilon_{J}\bb 1)+({\bb \Omega}+\epsilon_{J}\bb 1)\bb J(n)-2\gamma  {\bb\Omega}\big],
\end{equation}
with the initial condition $\bb J(n=0)=0$.
For simplicity, we fix $\gamma=1$ without loss of generality, as it only accounts for a global rescaling of the coupling matrix.  The dynamical equation Eq. \eqref{app:eqJ} can be rewritten as a fixed-point equation:
$$
\bb J(n+1)=  G_{\epsilon}(\bb J(n)),
$$
with
$$
 G_{\epsilon}(\bb J) = \bb J -\epsilon (\bb J ({\bb \Omega}+\epsilon _J \bb 1)+ ({\bb \Omega}+\epsilon _J \bb 1)\bb J-2\gamma {\bb \Omega}).
$$
Given $\bb J$ and $\bb K$ two $N\times N$ matrices, we have
\begin{equation*}
	\begin{split}
		 G_{\epsilon}(\bb J)- G_{\epsilon}(\bb K)= \frac12 (\bb J -\bb K)(\bb 1 -2\epsilon  ({\bb \Omega}+\epsilon _J \bb 1) )+ \frac12(\bb 1 -2\epsilon  ({\bb \Omega}+\epsilon _J \bb 1) ) (\bb J -\bb K).
	\end{split}
\end{equation*}
Taking the (operator) norm of both sides, we have
$$
\lVert  G_{\epsilon}(\bb J)- G_{\epsilon}(\bb K)\lVert \le \lVert \bb J-\bb K\lVert \cdot \lVert \bb 1 -2\epsilon  ({\bb \Omega}+\epsilon _J \bb 1) \lVert= \lVert \bb J-\bb K\lVert(1-2\epsilon (\lambda_1+\epsilon_J)),
$$
where $\lambda_1$ is the largest eigenvalue of ${\bb\Omega}$. This means that the function $G_\epsilon (\bb J)$ is a contraction map, provided that
\begin{equation}\label{dtb}
	\epsilon\leq\frac{1}{2(\epsilon_{J}+\lambda_{1})}.
\end{equation}
Thus, exploiting Banach Fixed Point Theorem, the algorithm converges to the solution of the system. Further, by Gershgorin's theorem,
we can get the following bound
\begin{equation}
	\lambda_1 \leq\sum_{ij}| \Omega_{ij}|.
\end{equation}
Thus, for any $\epsilon$ such that
\begin{equation}\label{discstep}
	\epsilon \le\frac{1}{2(\epsilon_{J}+\sum_{ij}| \Omega_{ij}|)},
\end{equation}
the convergence requirement \eqref{dtb} is trivially satisfied.
\par\medskip

\section{{Methods}\label{metodi}}
{
This appendix is devoted to a detailed description of numerical experiments.
We recall that our experiments encompass a training phase, in which we train the coupling matrix of the network with a training set, and a reconstruction phase, in which we perform a sequential dynamics on the neural configuration starting from an item belonging to a test set. Training and test sets are described in Sec.~\ref{sec:dataset} and Sec.~\ref{sec:numerics}: synthetic (Rademacher ground patterns) and benchmark (MNIST and Fashion-MNIST) datasets are considered.
To facilitate the reproducibility of the research, here we present
algorithms and pseudocodes of the numerical experiments (Subsec.~\ref{sec:B1}) along with the performance metrics to evaluate the quality of the final neuron states (Subsec.~\ref{sec:B2}); further evidence on the emergence of overfitting is also provided (Subsec.~\ref{app:num_overfitting}).
Moreover, we point out that all the simulations were carried out using the high-performance, dynamic programming language Julia and were run on a personal computer with an Intel Core i7 processor.}

\subsection{{Training design and parameters assignment}} \label{sec:B1}
{
In the training phase we find the expression of the coupling matrix $\boldsymbol J$ and of the external field $\boldsymbol h$ which minimize the loss function $\mathcal{L}_{\boldsymbol \xi}(\boldsymbol J,\boldsymbol h)$ given in equation \eqref{eq:loss_1} via gradient descent method. The parameters appearing in the loss function are summarized and described in Tab.~\ref{tab1}. Specifically, $N$ is the size of the input and is equal to the length of the training patterns, $P$ is the size of the training set (or total number of training patterns), $\epsilon_J$ is the regularization constant for the coupling matrix, $\epsilon_h$ is the regularization constant for the field $\bb h$ and, finally, $t$ is the training time.
In the simulations with structured datasets such as MNIST and Fashion-MNIST, since we want to reconstruct the original dataset and not its centered version, we preserve the field $\boldsymbol{h}$ in the dynamics and, for the sake of convenience and simplicity we  set $\epsilon_h=0$ in all the simulations.
The pseudo-code presented in Algorithm~\ref{training} highlights the steps followed to train the model.
}

{
If we run Algorithm~\ref{training} up to convergence, the resulting interaction matrix recovers the dreaming kernel, denoted as $\bb J^{(D)}$ and reported in Eq.~\eqref{eq:JD}, as long as we set $\epsilon_J = t_d^{-1}$ and $\gamma =1$. Thus, if we are interested in the regularized, full-trained model, we can directly pose $\bb J = \bb J^{(D)}$ without the need of running the training procedure.
If we set $\epsilon_{\boldsymbol J}=\epsilon_{\boldsymbol h}=0$ and we stop the gradient descent at the training $t=t^{*}$ given in \eqref{eq:tstar}, then the resulting coupling matrix $\boldsymbol J(t^*)$ is the closest possible to the interaction matrix $\bb J^{(D)}$.\\
\begin{table}
\begin{center}
{
\begin{tabular}{ |c|c| } 
\hline
Parameter & Description \\
\hline
$N$ & size of the input  \\ 
$P$ & size of the training set \\ 
$\epsilon_{J}$ & $L_2$-regularization term for $\boldsymbol J$ \\
$\epsilon_{h}$ & $L_2$-regularization term for $\boldsymbol h$ \\
$t$ & training time \\
\hline
\end{tabular}
\caption{\label{tab1}{Training parameters.}}}
\end{center}
\end{table}
}
\begin{algorithm}[tb]
\begin{algocolor}
{
\caption{ Training of the coupling matrix $\boldsymbol{J}$}\label{training}
\hspace*{\algorithmicindent} \textbf{Input}: Training set $\bb \xi = \{\boldsymbol \xi^\mu\}_{\mu=1}^P \in \{ -1, +1\}^{N \times P}$, stopping time $t$, regularizator $\epsilon_J$\\
\hspace*{\algorithmicindent} Settings for $t$ and $\epsilon_J$:\\
\hspace*{\algorithmicindent}\hspace*{\algorithmicindent}  $L_2$-regularization: $t=\infty$ and $\epsilon_J>0$\\
\hspace*{\algorithmicindent}\hspace*{\algorithmicindent} Early-stop regularization: $t = t^*$ as given by \eqref{eq:tstar}  and $\epsilon_J=0$\\
\hspace*{\algorithmicindent}\textbf{Output}: $\boldsymbol J (t),\boldsymbol h (t)$ 
\begin{algorithmic}[1]
\State $\boldsymbol{J(0)}=\bb 0_{N,N}$ inizialization of the coupling matrix 
\State $\hat{\boldsymbol{\xi}^\mu}=\boldsymbol\xi^\mu-\frac{1} {P} \sum_{\mu=1}^P\boldsymbol{\xi}^\mu$ pattern centering
\State $\hat{\boldsymbol{\Omega}}=\frac{1}{P}( \hat{\boldsymbol{\xi}} \cdot \hat{\boldsymbol{\xi}}^{T})$ computation of the Hebbian kernel 
\State $\Delta_t=\frac{1}{2}[\epsilon_J+\sum_{i,j=1}^N|\hat\Omega_{ij}|]^{-1}$ discretization time as given by \eqref{discstep}
\State $iters=\lc\frac{t}{\Delta_t}\rc$ number of iterations to be run
\State n=0
\Repeat 
\State $\boldsymbol{J}=\boldsymbol{J}-\Delta_t\left(2\epsilon_{J}\boldsymbol J +\boldsymbol J^{\prime}\hat{\boldsymbol{\Omega}}+\hat{\boldsymbol{\Omega}}^{\prime}\boldsymbol J-2\gamma\hat{\boldsymbol{\Omega}}\right)$
\State $n=n+1$
\Until{$n=iters$ or $\bb J$ has reached a fixed point.}\\
\Return $\boldsymbol J, \boldsymbol h=(\gamma\bb 1- \boldsymbol{J})\frac{1}{P}\sum_{\mu=1}^P\boldsymbol{\xi^{\mu}}$
\end{algorithmic}  }
\end{algocolor}
\end{algorithm}
{
}

\subsection{{Formulation of the performance metrics}} \label{sec:B2}
{
Once the training is over and we have the desired expression of the coupling matrix $\boldsymbol J$, the retrieval capabilities of the machine are investigated. The initial configuration $\bb \sigma^{(0)}$ is taken as a corrupted version of one of the training patterns (with a fraction of flipped entries w.r.t. the reference configuration) or as an item of a test set (whose elements are statistically analogous to the training patterns, but were not involved in the training procedure); in general, we denote with $\mathcal Q$ the sample of initial configurations. Then, the system relaxes according to the evolution rule introduced in Eq.~\eqref{eq:zeronoise} and reported hereafter in a discrete-time notation
\begin{equation}
	\sigma_{i}^{(n+1)}=\text{sign}(\sum_{j}J_{ij}\sigma_{j}^{(n)}+h_{i}\sigma_i^{(n)}).
\end{equation} 
The pseudo-code for the dynamics can be found in Algorithm \ref{dyn}. After the relaxation towards an equilibrium configuration $\bb \sigma^{(\infty)}$, we check its proximity to a  specific configurations $\bb \xi^*$ by exploiting the normalized Hamming distance as a performance metrics. The latter is the number of misaligned entries between two configurations $\bb \sigma^1,\bb \sigma^2\in \{-1,1\}^N$ divided by $N$, that is
$$d(\boldsymbol{\sigma}^{1},\boldsymbol{\sigma}^{2}):=\frac{1}{2N}\sum_{i=1}^N|\sigma_i^{1}-\sigma_i^{2}|.$$}

{Then, we evaluate $d(\bb\sigma^{(\infty)},\bb\xi^*)$, which depends on the initial configuration, due to the fact that $\bb\sigma^{(\infty)}$ depends on $\bb \sigma^{(0)}$. Next, we average $d(\bb\sigma^{(\infty)},\bb\xi^*)$ over the $|\mathcal Q|$ different realizations of the initial condition; when  $\bb \sigma^{(0)}$ is meant as an item of the test set, this operation corresponds to a batch average. This way, we get the average distance, defined as
}
\begin{equation}
	\bar d_{\bb\xi^*}=\frac{1}{|\mathcal Q|}\sum_{\bb\sigma^{(0)} \in \mathcal Q} d(\bb\sigma^{(\infty)}(\bb\sigma^{(0)}), \bb\xi^*),
\end{equation}
{
that is expected to depend on the reference point $\bb\xi^*$ and on the system parameters.}

{Finally, we notice that $d(\bb\sigma^{(\infty)},\bb\xi^*)$ is nothing but the (normalized) absolute error made by the machine outputting $\bb\sigma^{(\infty)}$, when asked to reconstruct $\bb\xi^*$.}

\begin{algorithm}[tb]
\begin{algocolor}
{
\caption{ Sequential dynamic}\label{dyn}
 \hspace*{\algorithmicindent} \textbf{Input}: Couplings $\boldsymbol J\in \mathbb{R}^{N\times N}$, external fields $\bb h \in \mathbb{R}^N$, input $\boldsymbol{\sigma}^{(0)} \in \{-1,1\}^{N}$
 \\
 \hspace*{\algorithmicindent} \textbf{Output}: Final neural configuration $\boldsymbol\sigma^{(\infty)}$
\begin{algorithmic}[1]
\State Remove the diagonal terms from $\boldsymbol J$
\State Set $n=0$
\Repeat 
\State sample a random integer $i$ uniformly in the set $\{1,2,\dots,N\}$
\State update the $i$-th spin $\sigma_{i}$ according to $\sigma_{i}^{(n+1)}=\text{sign}( \sum_{j=1}^N J_{ij}\sigma_{j}^{(n)}+h_i\sigma_i^{(n)})$
\State $n=n+1$
\Until{$\boldsymbol \sigma$ has reached a fixed point.}
\end{algorithmic} } 
\end{algocolor}
\end{algorithm}

\subsection{\label{app:num_overfitting}Numerical signatures of overfitting}
\begin{figure}[t!]
\centering
 \includegraphics[width=0.8\textwidth]{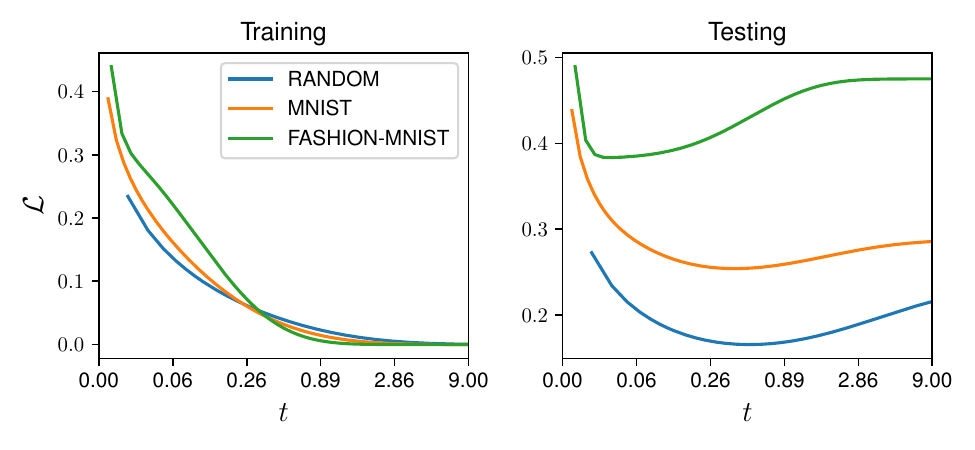}\\
 \caption{{\bfseries Training and test losses as a function of the training time.} The figure shows the comparison between the training loss function and the validation ones for synthetic dataset (blue curve), MNIST (orange curve) and Fashion-MNIST (green curve) for a training procedure without regularization term.}\label{fig:train_vs_val_losses}
 \end{figure}

In this section, we will give more numerical indications of overfitting emergence for the model we are dealing with.
Specifically, we check that overfitting can take place in the non-regularized training procedure. To do this, we follow the evolution of the following learning and validation loss functions over training time, in particular the loss functions have the following structure
{
\begin{equation}\label{losses}
    	\mathcal L_{\boldsymbol\xi} (\boldsymbol J (t),\boldsymbol h (t))=\frac1{2P} \sum_{i,\mu}(\sum_j J_{ij}(t) \xi^\mu_j +h_i(t)- \xi^\mu_i )^2+ \frac1{2P} \sum_{j,\mu}(\sum_i J_{ij} (t) \xi^\mu_i +h_j (t) -\xi^\mu_j )^2,
\end{equation}
where
\begin{equation}
     h_i(t) =\frac{1}{P} \sum_{\mu=1}^P( \xi_i^\mu-\sum_{j=1}^N J_{ij}(t)\xi_j^\mu)
\end{equation}
In the following numerical simulations, $J_{ij}(t)$ evolves according to Algorithm \ref{training} with parameters $\epsilon_J$.
Equation \eqref{losses} is the loss given in Eq. \eqref{eq:loss_1} of the main text with $\epsilon_J=\epsilon_h=0,\gamma=1$.
The  learning and validation loss functions are obtained by substituting into the previous equations $\{\boldsymbol \xi\}^P_{\mu=1}$ with the features of the training and test dataset respectively.} For the synthetic random dataset and MNIST and Fashion-MNIST cases we found that, even if the training loss goes to zero, the validation ones do exhibit a global minimum at finite training time, and then is start to increase, thus signalizing a worsening in generalization performances. These results are reported in Fig. \ref{fig:train_vs_val_losses}.

\section{A faster way to compute the early-stopping time}\label{sec:approxtstar}
\begin{figure}[tb]
	\centering
	\includegraphics[width=\textwidth]{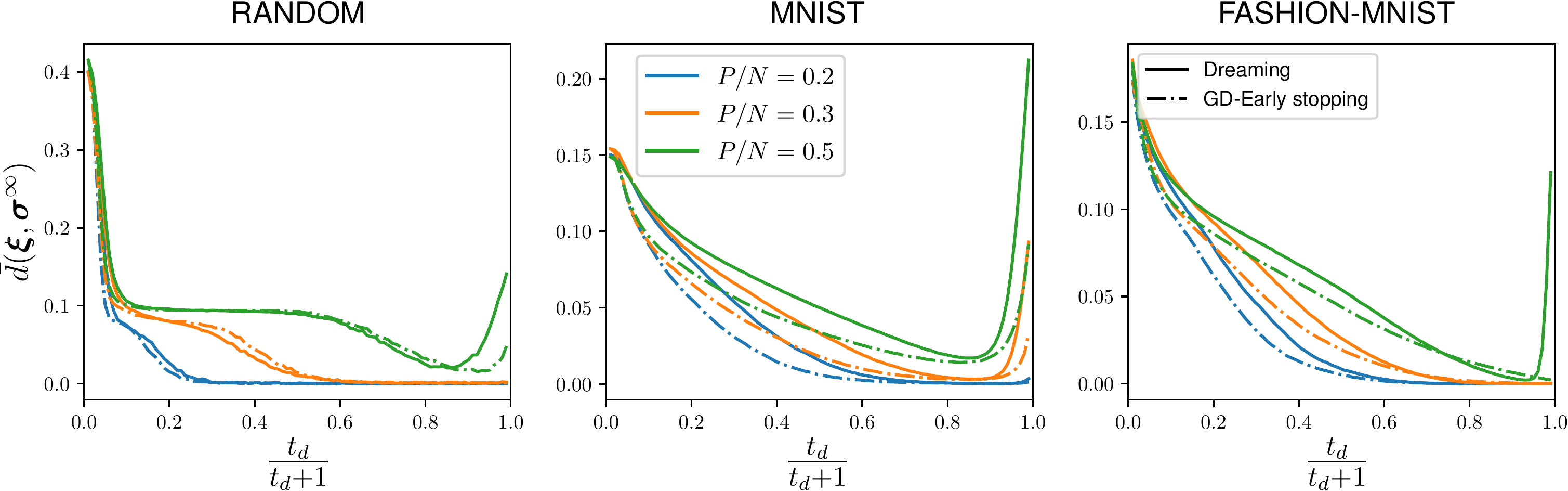}
	\caption{{\bfseries Comparison between early-stopping and dreaming kernel with approximated time.} The three plots show the comparison between the retrieval performances of the Dreaming kernel and the early-stopped training procedure. The content is perfectly specular to Fig. \ref{fig:Jd_vs_Jes}, with the only exception that the early-stopping time here is computed with the first-order approximation \eqref{eq:approxtstar}.}\label{fig:approxtstar}
\end{figure}

The estimate of the early-stopping time reported in Eq. \eqref{eq:tstar} is based on the computation of the empirical spectral distribution $\rho_E$ of ${\bb\Omega}$, which is a $N\times N$ matrix. For high-dimension datasets, this can constitute a bottleneck in the training procedure, so operative criteria needs to be provided. To do this, we can Taylor expand the quantity {$\delta(\lambda,t,\epsilon_J)$} around the average eigenvalue of the empirical distribution, i.e.
{$$
\delta(\lambda,t,\epsilon_J)= \sum_{k=0}^\infty \frac1{k!} \delta(\bar \lambda,t,\epsilon_J)^{(k)}(\lambda-\bar\lambda)^k,
$$}
thus
{
$$
\bar\delta(t,\epsilon_{J})=\sum_{k=0}^{\infty}\frac{1}{k!}  \delta(\bar \lambda,t,\epsilon_J)^{(k)}\int d\lambda \left(\lambda-\bar\lambda\right)^{k}\rho_E(\lambda)= \sum_{k=0}^\infty \frac1{k!}\delta(\bar \lambda,t,\epsilon_J)^{(k)}\frac1N \mathbb E\text{Tr} \,({\bb \Omega} -\bar\lambda \bb 1)^k.
$$}
Stopping at the first order in $\lambda-\bar\lambda$ and then solving the minimization problem, we are left with the prescription 
\begin{equation}\label{eq:approxtstar}
	t^{*}(t_d)\approx\frac{1}{2\frac1N \mE \text{Tr}\, {\bb\Omega}}\log\Big(1+t_d\frac1N \mE \text{Tr}\, {\bb\Omega}\Big).
\end{equation}
The results are reported in Fig.~\ref{fig:approxtstar}, which shows again a substantial agreement between the early-stopping procedure and the Dreaming kernel scenario.
Notice that this criterion precisely accounts for the correct logarithmic dependence of the early-stopping time w.r.t. the dreaming time $t_d$. Clearly, this prescription can provide, in the general case, a rough estimate of the early-stopping time. In that case, one can decide also to work out sub-leading orders in $\lambda -\bar\lambda$: regardless of the order at which the computation is performed, the numerical estimate of $t^*$ is based on the computation of low-order moments of the (modified) Hebbian kernel $\hat{\bb\Omega}$, which is far easier than to compute the whole empirical spectral distribution.

\section{Details on spurious states} \label{app: spurious}
In this Appendix, we report some details about spurious configurations of training examples in the synthetic random dataset. We recall that $\bb \xi^{\mu,A}= \bb \chi^{\mu,A}\bb \zeta^{\mu}$ (with entry-wise multiplication), with the $\chi^{\mu,A}_i$ variables extracted as
$$
\textrm{Prob}[\chi ^{\mu,A}_i =\pm1 ]= \frac{1\pm r}{2}.
$$
Relevant spurious configurations in this setup are of Hopfield-type, so that we can consider combination of the form
\begin{equation}
	{\bb \xi}^\mu_L= \text{sgn} \Big (\sum_{l=1}^L {\bb \xi^{\mu,A_l}}\Big).
\end{equation}
The correlation of this new configuration with the ground-pattern $\bb\zeta ^\mu$ is
$$
c_L = \frac1N \sum_i {\xi}^\mu_{L,i}\zeta_i^\mu=\frac1N\sum_i \text{sgn}\big( \sum_{l=1}^L  \xi^{\mu,A_l}_i\zeta^\mu_i\big)=\frac1N\sum_i \text{sgn}\big( \sum_{l=1}^L  \chi^{\mu,A_l}_i\big).
$$
For large $L$, the random variable in the sign function is, by central limit theorem (CLT), Gaussian distributed, with mean $r$ and variance $\sqrt{(1-r^2)/L}$, thus
$$
c_L = \frac1N\sum_i \text{sgn}\big(1+\sqrt{\frac{1-r^2}{Lr^2}}z_i\big),
$$
with $z\sim \mathcal N(0,1)$, where we also used $r>0$. The argument of the sign function is positive with probability
$$
p = P\big(z_i \ge -\sqrt {\frac{L r^2}{1-r^2}}\big)= 1-\frac12 \text{erfc}\Big[\sqrt{\frac{Lr^2}{2(1-r^2)}}\Big].
$$
Thus, the quantity $c_L$ is nothing but a random walk of unitary steps with probability $p$ to jump on the right. In the large $N$ limit, we thus have
$$
c_L(r)= \frac1N\sum_i \text{sgn}\big(1+\sqrt{\frac{1-r^2}{Lr^2}}z_i\big)\approx 2p-1= \text{erf}\Big[\sqrt{\frac{Lr^2}{2(1-r^2)}}\Big].
$$
which yields
	$$
	c_L(r) > r\quad \Rightarrow\quad L > 2\frac{[\text{erf}^{-1}(r)]^2(1-r^2)}{r^2}.
	$$
	Now, the r.h.s. of the last inequality ranges in $[0, \pi/2]$, while the approximated expression for $c_L(r)$ is obtained under the CLT, requiring $L \gg 1$, which is therefore sufficient for obtaining $c_L(r) > r$. \red{For fixed $L$ there are $\binom{M}{L}$ possible configurations of the form \eqref{eq:spurious}. Just to give an example of the magnitudes here involved, for $M=50$ and $r=0.6$, we have 1225 spurious combinations of $L=2$ examples displaying a correlation with ground-feature $c_2 \approx 0.8884$, we have $\sim 2\cdot 10^6$ combinations with $L=5$ displaying a correlation $c_5\approx 0.9993$, and so on.}


\bibliographystyle{unsrt}

\end{document}